\documentclass{elsart}

\usepackage{amsmath}
\usepackage{graphicx}
\usepackage{stmaryrd}
\usepackage{amssymb}
\usepackage{color}
\usepackage{multirow}
\usepackage{lineno}
\usepackage{algorithm}
\usepackage{algorithmic}
\usepackage{lscape} 
\usepackage{subcaption}

\usepackage{url}

\begin{document}

\begin{frontmatter}
\title{Population-based Gradient Descent Weight Learning for Graph Coloring Problems}

\author[Angers]{Olivier Goudet},
\ead{olivier.goudet@univ-angers.fr}
\author[Angers]{B\'eatrice Duval},
\ead{beatrice.duval@univ-angers.fr}
\author[Angers,IUF]{Jin-Kao Hao\corauthref{cor}}
\ead{jin-kao.hao@univ-angers.fr}
\corauth[cor]{Corresponding author.}  
\address[Angers]{LERIA, Universit$\acute{e}$ d'Angers, 2 Boulevard Lavoisier, 49045 Angers, France}
\address[IUF]{Institut Universitaire de France, 1 Rue Descartes, 75231  Paris, France}


\maketitle

\begin{abstract}
Graph coloring involves assigning colors to the vertices of a graph such that two vertices linked by an edge receive different colors. Graph coloring problems are general models that are very useful to formulate many relevant applications and, however, are computationally difficult. In this work, a general population-based weight learning framework for solving graph coloring problems is presented. Unlike existing methods for graph coloring that are specific to the considered problem, the presented work targets a generic objective by introducing a unified method that can be applied to different graph coloring problems. This work distinguishes itself by its solving approach that formulates the search of a solution as a continuous weight tensor optimization problem and takes advantage of a gradient descent method computed in parallel on graphics processing units. The proposed approach is also characterized by its general global loss function that can easily be adapted to different graph coloring problems. The usefulness of the proposed approach is demonstrated by applying it to solve two typical graph coloring problems and performing large computational studies on popular benchmarks. Improved best-known results (new upper bounds) are reported for several large graphs.  

\emph{Keywords}: Learning-based problem solving; heuristics; gradient descent; combinatorial search problems; graph coloring.
\end{abstract}
\end{frontmatter}


\section{Introduction}\label{Introduction}

Graph coloring problems are popular and general models that can be used to formulate numerous practical applications in various domains \cite{Lewis16}. Given an undirected graph $G=(V,E)$ with a set of vertices $V$ and a set of edges $E$, a legal coloring is a partition of vertex set $V$ into $k$ different color groups (also called independent sets) such that two vertices linked by an edge belong to different color groups. A legal coloring of $G$ can also be defined by a mapping that associates each vertex to a color such that two adjacent vertices must receive different colors (this is the so-called graph coloring constraint). Typically, the used colors or the color groups are represented by consecutive integers $1,\dots,k$. Consequently, the following statements are equivalent: vertex $v$ is assigned to (or receives) color $i$; $v$ is assigned to color group $i$.

Graph coloring problems generally involve finding a \textit{legal} coloring of a graph while considering some additional decision criteria and constraints. Specifically, the popular Graph Coloring Problem (GCP) is to determine the chromatic number of a graph, i.e., the smallest number of colors needed to reach a legal coloring of the graph. With an additional equity constraint, the Equitable graph Coloring problem (ECP) \cite{Meyer1973} requires finding a legal coloring with the minimum number of colors while the sizes of the color groups differ by at most one (i.e., the color groups are almost of the same size). When the number of colors $k$ is given, the Graph $k$-coloring Problem ($k$-COL) is to find a legal coloring of a graph with the $k$ given colors. 

In terms of computational complexity, graph coloring problems including those mentioned above are NP-hard and thus computationally challenging. Given their theoretical and practical significance,  graph coloring problems have been studied very intensively in the literature. However, as the review on the GCP and the ECP of Section 2 shows, studies on graph coloring  are usually  specific to a problem and it is  difficult to generalize a method designed for a coloring problem to other coloring problems even if they are tightly related. 

This work aims to target  a more general objective for solving graph coloring problems by proposing a framework that can be applied to different graph coloring problems and more generally grouping problems where a set of items needs to be separated into different groups according to some given grouping constraints and objectives. The contributions of this work are summarized as follows.

First, a population-based gradient descent weight learning approach with a number of original features is proposed. A coloring problem is considered from the perspective of continuous optimization to benefit from the powerful gradient descent. Specifically, a weight matrix representation of a candidate coloring is used, where each entry of the matrix corresponds to a learned propensity of a vertex to receive a particular color. Then a gradient descent method is used to improve a population of candidate solutions (or individuals) by minimizing a global loss function. The global loss function is designed to effectively guide the gradient descent. To be informative, the global loss function aggregates different terms: a fitness term measuring the quality of a candidate solution, a penalization term aiming to discourage the individuals to repeat the same color assignment mistakes and a bonus term aiming to encourage the individuals to keep the shared correct assignments. To solve this problem, the tensor of candidate solutions is expressed as a tensor of continuous weight matrices.  Accelerated training with Graphics Processing Unit (GPU) is used to deal with multiple parallel solution evaluations. As the first study of using gradient descent based weight learning to solve graph coloring problems with tensor calculations, this work enriches the still limited toolkit of general solution methods for this important class of problems.

Secondly,  the usefulness of the proposed framework is demonstrated by applying it to two popular graph coloring problems: the GCP and the ECP. For both problems, extensive computational experiments are presented on well-known benchmark graphs and show the competitiveness of our approach compared to state-of-the-art algorithms. In particular, improved best results are presented for several large geometric graphs and large random graphs for the ECP.

Thirdly, this work shows the viability of formulating a discrete graph coloring problem as a real-valued weight learning problem. The competitive results on two challenging representative graph coloring problems invite more investigations of testing the proposed approach to other graph coloring problems and related grouping problems.

The rest of the paper is organized as follows. Section \ref{sec:related_work} reviews related heuristics for the GCP and the ECP. Section \ref{sec:proposed_approach} describes the proposed approach and its ingredients. Section \ref{sec:exp_analyzes} provides illustrative examples on the gradient descent learning. Section \ref{sec:experiments} reports computational results on well-known benchmark graphs and shows comparisons with the state-of-the-art. Section \ref{Discussion} discusses the applicability of the proposed framework to other graph problems. Section \ref{sec:conclusion} summarizes the contributions and presents perspectives for future work.

\section{Reference heuristics for the two coloring problems}
\label{sec:related_work}

The GCP and the ECP have been studied very intensively in the past. Both problems are known to be NP-hard in the general case (see \cite{garey1979computers} for the GCP and \cite{furmanczyk2005complexity} for the ECP). Thus, assuming that $N\neq NP$, no algorithm can solve these problems exactly in polynomial time except for specific cases. In practice, there are graphs with some 250 vertices that cannot be solved optimally by any existing exact algorithm (see \cite{malaguti2011exact} for the GCP and \cite{mendez2015dsatur} for the ECP).

Therefore, to handle large graphs, heuristics are typically used to solve these problems approximately. Generally, heuristics proceed by successively solving the $k$-coloring problem with decreasing values of $k$ \cite{galinier2013recent} and the smallest possible $k$ gives an upper bound of the optimum.

A thorough review of existing heuristics for the GCP and the ECP is out of the scope of this work. Instead,  a brief description of the main heuristics including in particular the best-performing heuristics is presented below. For a comprehensive review of these problems, the reader is referred to the following references for the GCP: local search heuristics \cite{galinier2006survey}, exact algorithms \cite{malaguti2010survey} and new developments on heuristics till 2013 \cite{galinier2013recent}. For the ECP, both \cite{SunHao2020} and \cite{wang2018tabu} provide a brief review of main heuristic algorithms.


\subsection{Heuristics for the graph coloring problem \label{sec:related_work_gcp}}

There are different ways of classifying  for the GCP. Following \cite{galinier2013recent}, these heuristics can be classified in three main categories: local search based methods, hybrid methods combining population-based genetic search and local search, and large independent set extraction. 

Local search heuristics iteratively improve the current solution by local modifications (typically, change the color of a vertex) to optimize a cost function, which typically corresponds to the number of conflicting edges (i.e., edges whose endpoints being assigned the same color). To be effective, a local search heuristics usually integrate different mechanisms to escape local optima traps. One of the most popular local search heuristics for the GCP is the \textit{TabuCol} algorithm \cite{hertz1987using}. \textit{TabuCol} iteratively makes transitions from the current solution to a new (neighbor) solution by changing the color of an endpoint of a conflicting edge according to a specific rule. In order to avoid search cycling, it uses a special memory (called tabu list) to prevent a vertex from receiving its former color. Despite its simplicity, \textit{TabuCol} performs remarkably well on graphs of reasonable sizes (with less than 500 vertices). This algorithm was largely used as the key local optimization component of several state-of-the-art hybrid algorithms (see below). Recently, \textit{TabuCol} has also been adopted as the local optimizer of the probability based learning local search method (PLSCOL) \cite{zhou2018improving}. Finally, this single trajectory local search approach has been extended to multiple trajectory local search, which is exemplified by the Quantum Annealing algorithm (QA) \cite{titiloye2011graph}. In QA, $n$ candidate solutions are optimized by a simulating annealing algorithm, while the interactions between the solutions occur through a specific pairwise attraction process, encouraging the solutions to become similar to their neighbors and improving the spreading of good solution features among new solutions.

The second category of methods includes hybrid algorithms, in particular, based on the \textit{memetic} framework \cite{MemeticBook2012}. These hybrid algorithms combine the benefits of local search for intensification with a population of high-quality solutions offering diversification possibilities. Different solutions evolve separately using a local search algorithm, such as the above-mentioned \textit{TabuCol} algorithm, and a crossover operator is used to create new candidate solutions from existing solutions. One of the most popular crossover operators for the GCP is the Greedy Partition Crossover (GPX) introduced in the hybrid evolutionary algorithm (HEA) \cite{galinier1999hybrid}. HEA follows the idea of grouping genetic algorithms \cite{falkenauer1998genetic} and produces offspring by choosing alternatively the largest color class in the parent solutions. The MMT algorithm \cite{malaguti2008metaheuristic} completes a first phase of optimization with a \textit{memetic} algorithm by a second phase of optimization based on the set covering formulation of the problem and using the best independent sets found so far. An extension of the HEA algorithm called MACOL (denoted as MA in this paper) was proposed in \cite{lu2010memetic}. It introduces a new adaptive multi-parent grouping crossover (AMPaX) and a new distance-and-quality based replacement mechanism used to maintain the diversity of the solutions in the population. Recently, a very powerful variation of the HEA algorithm, called hybrid evolutionary algorithm in duet (HEAD), was proposed in \cite{moalic2018variations}, which relies on a population of only two solutions. In order to prevent a premature convergence, HEAD introduces an innovative diversification strategy based on an archive of elite solutions. The idea is to record the best solutions (elite solutions) obtained in previous few generations and to reintroduce them in the population when the two solutions becomes too similar.  
  
The third category of approaches is based on extracting large independent sets in order to copy with the difficulty of coloring very large graphs \cite{wu2012coloring}. With this approach, an independent set algorithm is used to identify as many possible independent sets of large sizes and a coloring algorithm is employed to color the residual graphs (whose size is largely reduced). These methods are particularly effective to color large graphs with at least 1000 vertices. 

One notices that the above algorithms cover the current best-known results on the benchmark instances used in the experimental section of this work. However, no single algorithm is able to reach the best-known results for these hard instances. Moreover, it is worth mentioning that most current best-known results have not been updated for a long time, indicating the high difficulty of further improving these results.

Finally, it is worth mentioning that there are a number of other algorithms based on fashionable metaheuristics including ant colony optimization \cite{Ant2008}, particle swarm optimization \cite{PSO20152}, and other fanciful bio-inspired methods (flower pollination \cite{Flower2015}, firefly \cite{Firefly2017} and cuckoo optimization \cite{cuckoo2015}). However, the computational results reported by these algorithms (often on easy benchmark instances only) show that they are not competitive at all compared to the state-of-the-art coloring algorithms.

\subsection{Heuristics for the equitable graph coloring problem \label{sec:related_work_ecp}}

Most of the above-mentioned best performing methods for the GCP rely in one way or another on extraction of large independent sets or spreading of large building blocks between solutions. Thus, these methods cannot generally be applied as it stands for the ECP, because regrouping vertices in large sets may conflict with the equity constraint. Therefore different approaches have been proposed in the literature for the ECP.

One of the first method is the tabu search algorithm TabuEqCol \cite{diaz2014tabu}, which is an adaptation of the popular \textit{TabuCol} algorithm designed for the GCP \cite{hertz1987using} to the ECP. This algorithm was improved in \cite{lai2015backtracking} by the backtracking based iterated tabu search (BITS), which embeds a backtracking scheme under the iterated local search framework. These two algorithms only consider feasible solutions with respect to the equity constraint, which may restrict  the exploration of the space search too much.

A new class of methods relaxes the equity constraint and considers both equity-feasible and equity-infeasible solutions to facilitate the transition between visiting structurally different solutions. By enlarging their search spaces, these algorithms are able to improve on the results reported by the basic tabu search algorithm \cite{diaz2014tabu} and the backtracking based iterated tabu search algorithm \cite{lai2015backtracking}. The first work in this direction is the Feasible and Infeasible Search Algorithm (FISA) \cite{sun2017feasible}. The hybrid tabu search (HTS) algorithm \cite{wang2018tabu} is another important algorithm exploring both equity-feasible and equity-infeasible solutions and applying an additional novel cyclic exchange neighborhood. Finally, the latest memetic algorithm (MAECP) \cite{SunHao2020} is a population-based hybrid algorithm combining a backbone crossover operator, a two-phase feasible and infeasible local search and a quality-and-distance pool updating strategy. MAECP has reported excellent results on the set of existing benchmark instances in the literature. Among the algorithms, FISA, HTS and in particular MAECP hold the current best-known results on the benchmark instances used in the experiments of this work.

The above review indicates that the existing approaches for the GCP and the ECP are specific (i.e., they are specially designed for the given problem) and thus cannot be applied to other (even tightly related) coloring problems. For instance, the best performing methods for the GCP such as the multi-parent memetic algorihm \cite{lu2010memetic}, the two phase memetic algorithm \cite{malaguti2008metaheuristic} and the two individual memetic algorithm \cite{moalic2018variations} are based on specific crossover operators to combine large color classes. Such a strategy is meaningful for the GCP, but becomes inadequate for the ECP due to the equity constraint. As such, these approaches lack generality and are not readily applicable to other problems.

The current work aims to propose a more flexible and general framework for graph coloring problems, which is less dependent on the type of coloring problems or the type of graphs. This goal is achieved by formulating the computation of a solution of a coloring problem as a weight tensor optimization problem, which is solved by a first order gradient descent. As such, this approach can benefit from GPU accelerated training to deal with multiple parallel solution evaluations, ensuring a large examination of the given search space. As the computational results presented in Section \ref{sec:experiments} show, this unified approach, with slight modifications, is able to compete favorably with state-of-the-art algorithms for both the GCP and the ECP, leading to record breaking results for several large graphs for the ECP. Generally, this work narrows the gap between discrete optimization and gradient descent based continuous optimization, the latter being routinely used in particular in machine learning to train deep neural networks with billion of parameters.

\section{Population-based gradient descent learning for graph coloring} 
\label{sec:proposed_approach}

Given a graph $G = (V,E)$ with vertex set $V=\{v_i\}_{i=1}^n$ and edge set $E$, the considered problem is to partition $V$ into a minimum number of $k$ color groups $g_i$ $(i = 1,\dots,k)$, while satisfying some constraints. For the GCP, only the coloring constraint is imposed while for the ECP, the equity constraint must be additionally satisfied. To minimize the number of color groups, a typical approach is to solve a series of problems with different fixed $k$. For instance, the GCP can be approximated by solving a sequence of $k$-COL problems, i.e., finding $k$-colorings for decreasing $k$ values \cite{galinier2013recent} such that each $k$ leading to a legal coloring is an upper bound of the chromatic number of the graph. This approach is adopted to handle both the GCP and the ECP in this work.

\subsection{Tensor representation of a population of candidate solutions}
\label{Tensor representation of a population of candidate solutions}

A candidate solution for a graph coloring problem with $k$ color groups can be represented by a binary matrix $S=\{s_{i,j}\}$ in $\{0,1\}^{n\times k}$, with $s_{i,j}=1$  if the $i$-th vertex $v_i$ belongs to the color group $j$ and $s_{i,j}=0$ otherwise. Therefore, $\sum_{j=1}^k s_{i,j} = 1$, as one vertex belongs exactly to one color group. For the two graph coloring problems  considered in this work (GCP and ECP),  the constraints and the optimization objective of each problem are formulated as the minimization of a single fitness function $f:\{0,1\}^{n\times k} \rightarrow \mathbb{R}$. The goal of the problem is then to find a solution $S$ such that $f(S)$ is minimum. The definition of this function for the GCP and the ECP is provided in Section \ref{Step 2 - Fitness evaluation}.

In the proposed approach, a population of $D$ candidate solutions (individuals) will be considered and their fitness scores will be computed in parallel with tensor calculus on Compute Unified Device Architecture (CUDA) for GPU hardware. In order to do that, the $D$ matrices $S_d$ ($d \in \llbracket 1, D \rrbracket$) of the population are regrouped in a single three-dimensional tensor $\mathbf{S}=\{s_{d,i,j}\}$ in $\{0,1\}^{D \times n \times k }$. For a tensor $\mathbf{S}$ of $D$ candidate solutions, its \textit{global fitness function} is defined by the sum of all individual fitness $\sum_{d=1}^D f(S_d)$ and will be written as $f(\mathbf{S})$.

Hereafter, \textbf{bold} symbols are used to denote three-dimensional tensors and normal symbols to represent the underlying two-dimensional matrices, just like tensor $\mathbf{S}$ of $D$ candidate solutions and a single solution matrix $S$. 

\subsection{Weight formulation of a coloring problem}

Inspired by the works on reinforcement learning based local search \cite{zhou2016reinforcement} and ant colony optimization \cite{costa1997ants}, a real-valued weight matrix $W=\{w_{i,j}\}$ in $\mathbb{R}^{n \times k}$ is employed. It is composed of $n$ real numbers vectors $\mathrm{w_{i}}$ of size $k$  corresponding to a learned propensity of each vertex $v_i$ to receive a particular color. In order to build a discrete solution $S$ from $W$, each vertex $v_i$ will be assigned to the color group with the maximum weight in $\mathrm{w_{i}}$.

As a main idea, a real-valued weight tensor $\mathbf{W}=\{w_{d,i,j}\}$ in $\mathbb{R}^{D \times n \times k}$ of $D$ matrices $W_d$ is used to compute a tensor $\mathbf{S}$ of $D$ discrete coloring solutions. This is simply achieved by a function $g: \mathbb{R}^{D \times n \times k} \rightarrow \{0,1\}^{D \times n \times k }$  such  that $\mathbf{S} = g(\mathbf{W})$ (cf. Section \ref{step1} below). As $\mathbf{S}$ derives from $\mathbf{W}$, the given coloring problem becomes then the one of finding $\mathbf{W}$ such that $f \circ g (\mathbf{W})$ is minimum.

Our approach starts with a random weight tensor $\mathbf{W}$ where each entry is sampled independently from a normal distribution $\mathcal{N}(0,\sigma_0)$. Then, as explained below, a first order gradient descent is employed to update this weight tensor $\mathbf{W}$ with information learned during the search.

\subsection{General scheme}

The population-based gradient descent weight learning for graph coloring (or simply TensCol) proposed in this work relies on a population of candidate solutions represented as a weight tensor and performs tensor calculations in parallel with GPU by minimizing a flexible global loss function that encourages correction of wrong vertex-to-color assignments and consolidation of correct assignments of the solutions.

\begin{figure}[H]
    \centering
    \includegraphics[width=0.9\textwidth]{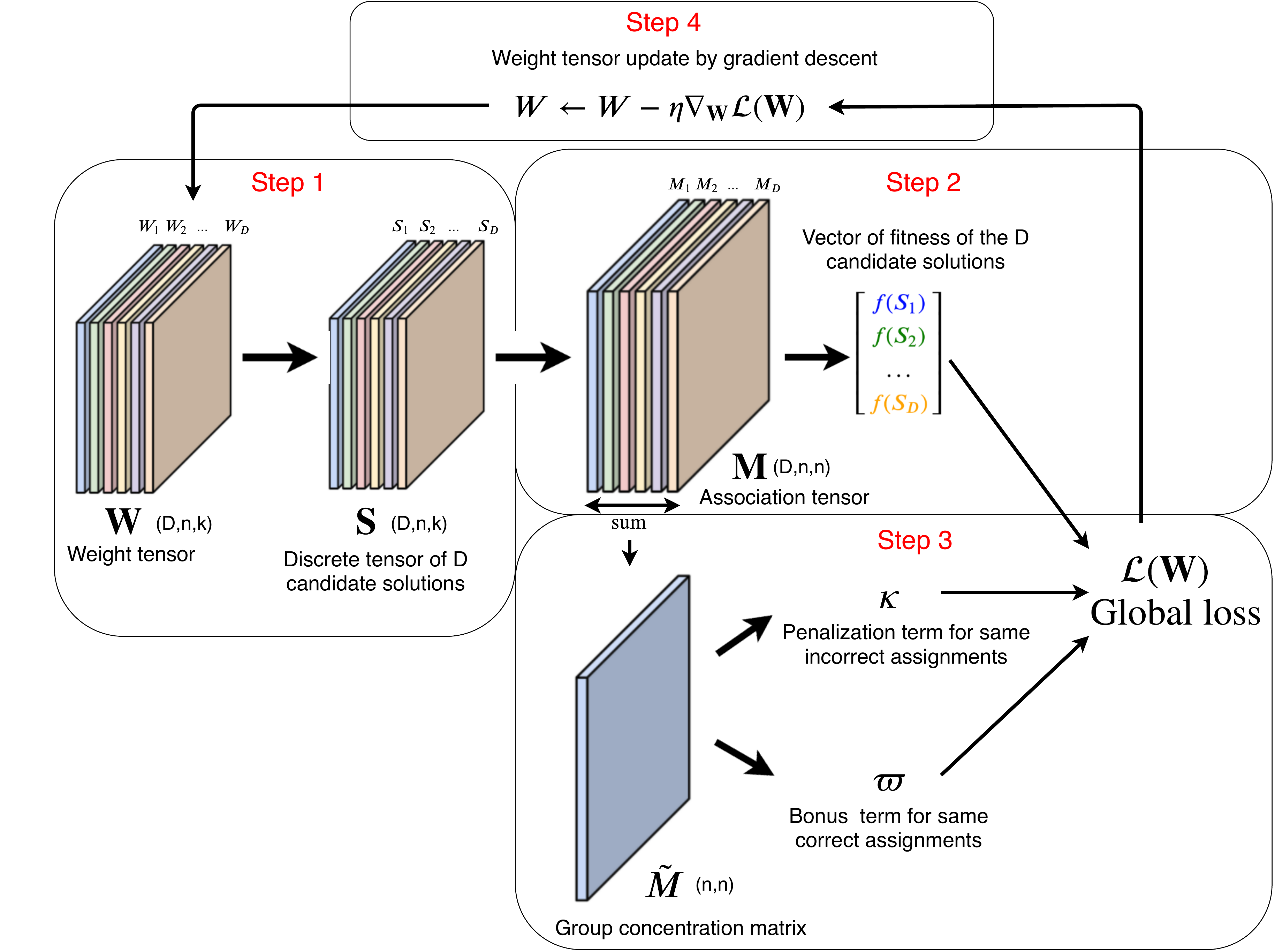}
    \caption{General scheme of the TensCol algorithm \label{General scheme}}
\end{figure}

From the perspective of algorithmic procedure, the proposed TensCol algorithm iteratively improves its population in four steps, as illustrated by Figure \ref{General scheme}. First, given the weight tensor $\mathbf{W}$,  the associated population of solutions  $\mathbf{S}$  (colorings) is computed (Step 1). From this population of solutions  $\mathbf{S}$, the vector of fitness $(f(S_1),f(S_2),...,f(S_D))$ is evaluated (Step 2). If a legal solution $S_d$ ($d \in \llbracket 1, D \rrbracket$) sothat $f(S_d)=0$ is found in the population, the algorithm stops. Otherwise a global loss term $\mathcal{L}$ (scalar) is computed (Step 3). $\mathcal{L}$ aggregates three terms: (i) $f(\mathbf{S})$ the global coloring fitness of the candidate solutions (see Section \ref{Step 2 - Fitness evaluation}), (ii) a penalization term $\kappa(\mathbf{S})$ for shared wrong color group assignments in the population (see Section \ref{penalization-term}) and  (iii) a bonus term $\varpi(\mathbf{S})$ for shared correct color group assignments (see Section \ref{bonus-term}). Finally, the gradient of the loss $\mathcal{L}$ with respect to the weight tensor $\mathbf{W}$ is evaluated  and then used to update the weights by first order gradient descent (Step 4). In the following subsections, each of these steps is described in detail.

\subsection{Step 1 - Vertex to color assignment \label{step1}}

\begin{figure}[H]
    \centering
    \includegraphics[width=1\textwidth]{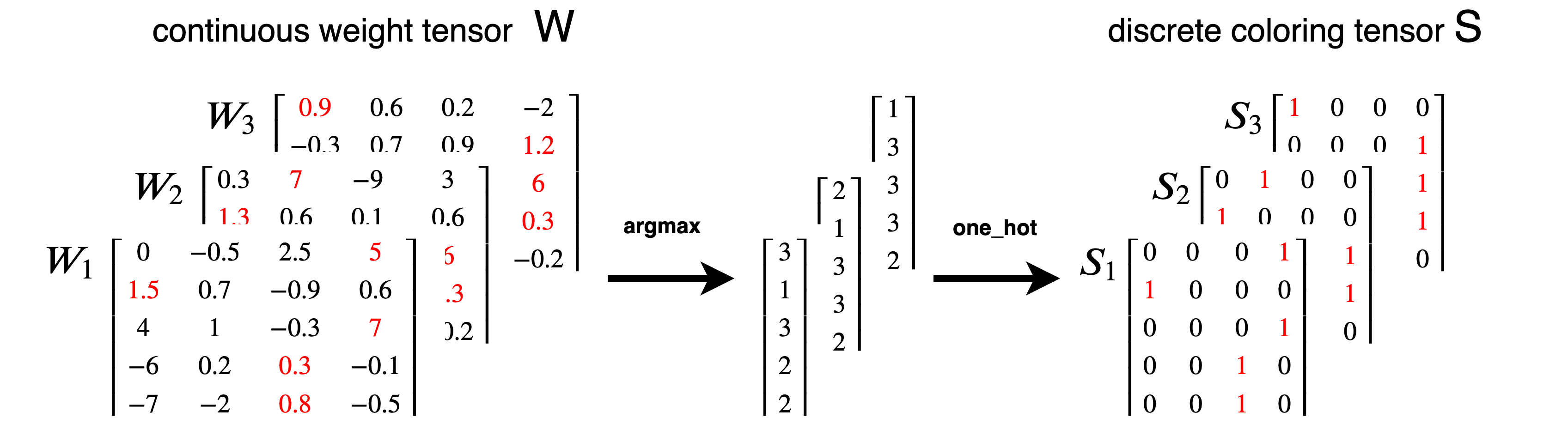}
    \caption{\label{fig:argmax} Example of vertex to color assignment $\mathbf{S}$ from a weight tensor $\mathbf{W}$, with $D=3$, $n=5$ and $k=4$.}
\end{figure}

At each iteration of the TensCol algorithm, the current weight tensor $\mathbf{W}$ is used to derive the associated tensor $\mathbf{S}$  of candidate solutions as follows. For $(d,i,j) \in \llbracket 1, D \rrbracket \times \llbracket 1, n \rrbracket \times \llbracket 1, k \rrbracket$, $s_{d,i,j} = 1$ if $j = \underset{l \in \{1,...,k\}}{\text{argmax}}\ w_{d,i,l}$ and $s_{d,i,j} = 0$ if $j \neq \underset{l \in \{1,...,k\}}{\text{argmax}}\ w_{d,i,l}$. This color group assignment for each vertex can be summarized as a single function  $g(\mathbf{W}) = \text{one\_hot}(\text{argmax}(\mathbf{W}))$ from $\mathbb{R}^{D \times n \times k}$ to $\{0,1\}^{D \times n\times k}$, where \textit{argmax} and \textit{one\_hot} operations are applied  along the last axis of the tensor (color axis). Figure \ref{fig:argmax} shows an example with $D=3$ individuals in the population, $n=5$ vertices and $k=4$ colors.

\subsection{Step 2 - Fitness evaluation}
\label{Step 2 - Fitness evaluation}

For a graph $G = (V,E)$ with $n$ vertices, let $A=\{a_{i,j}\}_{i,j=1...n}$  be its  adjacency  matrix sothat $\{v_i,v_j\} \in E$ if and only if $a_{i,j} = 1$. This symmetric matrix defines the coloring constraints to be respected by a legal coloring. The three-dimensional tensor $\mathbf{A}$ is used to represent the $D$ duplications of the adjacency matrix $A$.

Given  $\mathbf{S}$  the tensor of $D$ candidate solutions, $\mathbf{S}'$ denotes the transposed three-dimensional tensor obtained from $\mathbf{S}$ by swapping its second and third axis and  an \textit{association tensor} $\mathbf{M}$ is defined as $\mathbf{M} = \mathbf{S} \cdot \mathbf{S}'$, where $\cdot$ is the dot product between two tensors (sum product over the last axis of $\mathbf{S}$ and the second axis of $\mathbf{S}'$).

For $(d,i,j) \in \llbracket 1, D \rrbracket \times \llbracket 1, n \rrbracket \times \llbracket 1, n \rrbracket$, each entry of $\mathbf{M}$ is given by $m_{d,i,j} = \sum_{l=1}^k s_{d,i,l} s_{d,j,l}$. One notices that $m_{d, i,j} = 1$ if and only if the two vertices $v_i$ and $v_j$ in the solution $S_d$ are assigned the same color (i.e. belong to the same color group). Interestingly enough this \textit{association tensor} is independent by permutation of the columns in each candidate solution $S_d$ (permutation of the $k$ colors)\footnote{In the GCP and the ECP, two solutions $S$ and $S'$, such that $S'$ is equal to $S$ up to a permutation of its $k$ columns, are strictly equivalent in terms of fitness evaluation. It is very important to take into account this property in order to make  relevant comparisons of the candidate solutions in the population.}. 
Then, by performing the element-wise product between the tensors $\mathbf{A}$ and $\mathbf{M}$, the global \textit{conflict tensor} for the $D$ solutions is obtained  as $\mathbf{C} = \mathbf{A} \odot \mathbf{M}$, where $\odot$ corresponds to the Hadamar product. For $(d,i,j) \in \llbracket 1, D \rrbracket \times \llbracket 1, n \rrbracket \times \llbracket 1, n \rrbracket$, each entry of $\mathbf{C}$ is given by $c_{d,i,j} = a_{i,j}m_{d,i,j}$, and so $c_{d,i,j} = 1$ if and only if vertices $v_i$  and $v_j$ are \textit{in conflict} in the candidate solution $S_d$, i.e., they are linked by an edge ($a_{i,j}=1$) and are assigned to the same color group ($m_{d,i,j}=1$). As $c_{d,i,j}=c_{d,j,i}$, the total number of conflicts or the \textit{fitness} $f_{color}(S_d)$ of the candidate solution $S_d$ is then given by
 
\begin{equation} 
f_{color}(S_d) = \frac{1}{2}\sum_{i,j=1}^n c_{d,i,j}.
\label{eq:individual fitness}
\end{equation}

So the fitness vector containing $D$ fitness values (conflicts) for the whole population of solutions $\mathbf{S}$ is computed in parallel with a single tensor operation from $\mathbf{C}$ as $(f_{color}(S_1),..., f_{color}(S_d)) = \frac{1}{2} sum(\mathbf{C},(2,3))$ where $sum(\cdot,(2,3))$ is the summation operation along second and third axis. The \textit{global coloring fitness} for the tensor $\mathbf{S}$ of $D$ candidate solutions can be evaluated as

\begin{equation}
f_{color}(\mathbf{S}) = \frac{1}{2} sum(\mathbf{C}).
    \label{eq:global coloring fitness}
\end{equation}

For the classical GCP, any solution $S_d$ satisfying $f_{color}(S_d)=0$ is a legal coloring for the problem. As such, this global coloring fitness $f_{color}(\mathbf{S})$ is used as the global fitness function $f(\mathbf{S})$ to be minimized in the case of the GCP (see Section \ref{Tensor representation of a population of candidate solutions}): $f(\mathbf{S}) = f_{color}(\mathbf{S})$. 

For the ECP where the additional equity constraint is imposed, a supplementary equity fitness $f_{eq}(S_d)$ is introduced for each candidate solution $S_d$. Indeed, the equity constraint states that the number of vertices assigned to each of the $k$ color groups must be equal to $c_1=\lfloor \frac{n}{k} \rfloor$ or $c_2=\lfloor \frac{n}{k} \rfloor + 1$ where $\lfloor \cdot \rfloor$ denotes the integer part of a positive real number and $n$ is the number of vertices in ${G}$, with a particular case of $c_1=c_2$ when $n$ is divisible by $k$. To take into account the equity constraint, the additional \textit{equity fitness} $f_{eq}(S_d)$ is defined as follows:

\begin{equation}
f_{eq}(S_d) = \sum_{l=1}^k min(|\sum_{i=1}^{n}s_{d,i,l}-c_1|,|\sum_{i=1}^{n}s_{d,i,l}-c_2|).
\label{eq:equity_fitness_v0}
\end{equation}

It corresponds to the total number of surplus or deficit in terms of the number of vertices in the color groups with respect to the admissible numbers of vertices $c_1$ and $c_2$ in each group. A legal solution $S_d$ for the ECP must simultaneously satisfy $f_{color}(S_d)=0$ and $f_{eq}(S_d)=0$. Let $f_{eq}(\mathbf{S}) = \sum_{d=1}^D f_{eq}(S_d)$ be the \textit{global equity fitness} for the whole population $\mathbf{S}$ of $D$ candidate solutions. Then the global fitness function for the ECP (to be minimized) is given by $f(\mathbf{S}) = f_{color}(\mathbf{S}) + f_{eq}(\mathbf{S})$.

\subsection{Step 3 - Computing global loss with penalization and bonus}

Given a population $\mathbf{S}$ of $D$ solutions, the key idea is to introduce two dependency terms linking the $D$ solutions for two considerations: discourage that the solutions repeat the same mistakes of creating conflicts for the same pairs of vertices (penalization) and encourage the solutions to consolidate the correct group assignments (bonus).

For this purpose,  a summation of the association tensor $\mathbf{M}$ (see Section \ref{Step 2 - Fitness evaluation}) is  performed along the first axis.  The \textit{group concentration matrix} $\tilde{M}$ of size $n \times n$ is obtained:

\begin{equation}
    \tilde{M} = sum(\mathbf{M},1),
    \label{eq:tildeM}
\end{equation}

\noindent For $(i,j) \in  \llbracket 1, n \rrbracket \times \llbracket 1, n \rrbracket$, each entry  of $\tilde{M}$ is $\tilde{m}_{i,j}= \sum_{d=1}^D m_{d,i,j}$ which indicates the number of candidate solutions where vertices $v_i$ and $v_j$ are assigned to the same color group. The construction of this \textit{group concentration matrix} is independent by permutation of the $k$ colors (or $k$ groups) of each candidate solution $S_d$. 

\subsubsection{Penalization term for shared wrong color assignments \label{penalization-term}}

Using the \textit{group concentration matrix} $\tilde{M}$ above,  the penalization term $\kappa(\mathbf{S}) = \text{sum}(A \odot \tilde{M}^{\circ \alpha})$ of the population $\mathbf{S}$ is computed, where $\text{sum}(\cdot)$ corresponds to a sum of all matrix elements and  $^\circ$ designates element-wise power tensor calculation.

\begin{align}
\kappa(\mathbf{S})  &= \sum_{i,j=1}^n a_{i,j} \tilde{m}_{i,j}^{\alpha},
 \label{eq:diversity}
\end{align}

\noindent   where $\alpha$ is a parameter greater than 1 in order to penalize the conflicts shared by candidate solutions. By minimizing this term, the solutions are discouraged to make the same assignment mistakes for the same pair of vertices.

\begin{figure}[h]
    \centering
    \begin{subfigure}[t]{1\textwidth}
    \includegraphics[width=1\textwidth]{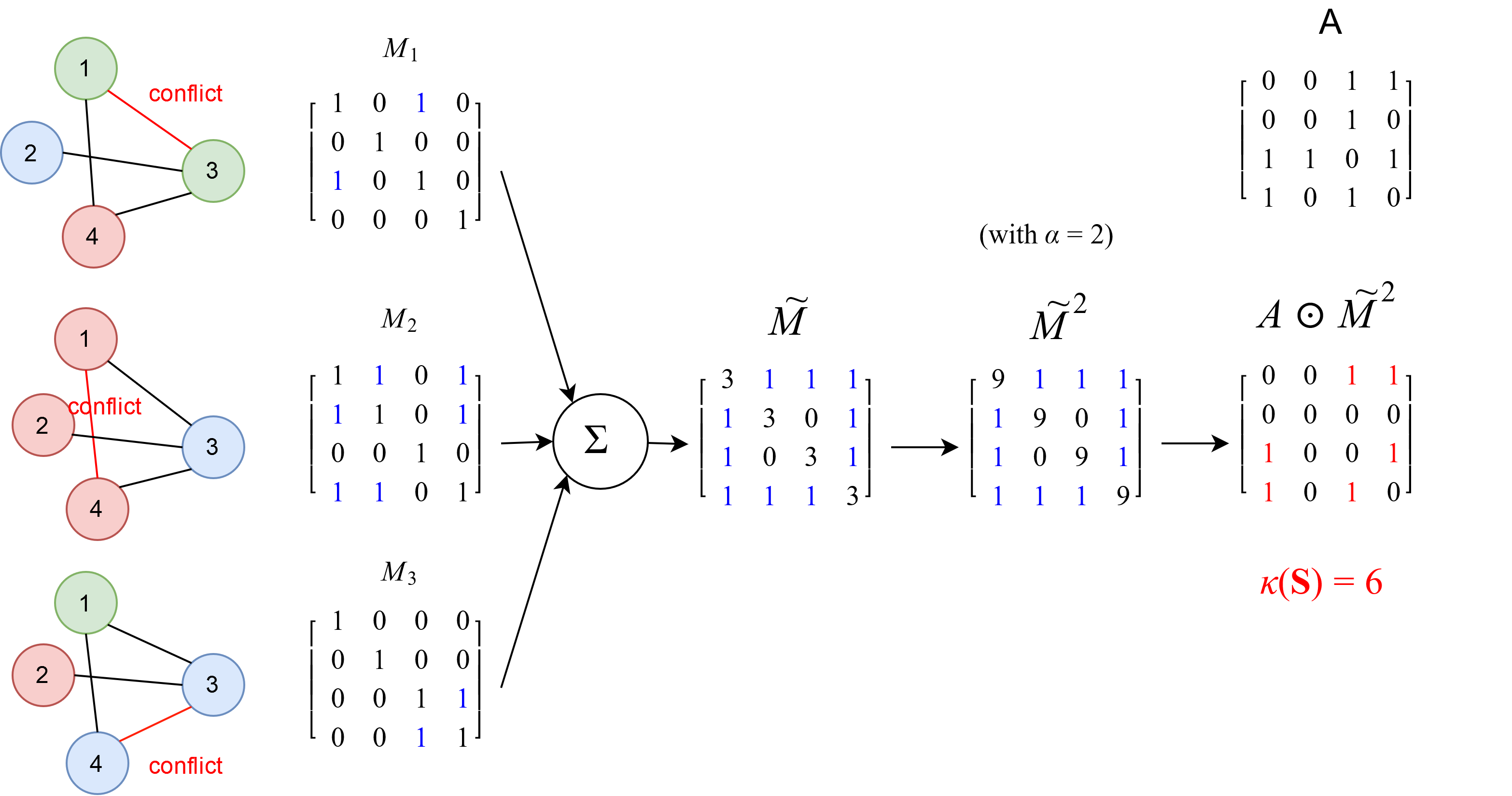}
    \caption{$\kappa(\mathbf{S})=6$ for a population of 3 candidate solutions with different conflicts.\label{fig:diff_conflict}}
    \end{subfigure}
    \begin{subfigure}[t]{1\textwidth}
    \includegraphics[width=1\textwidth]{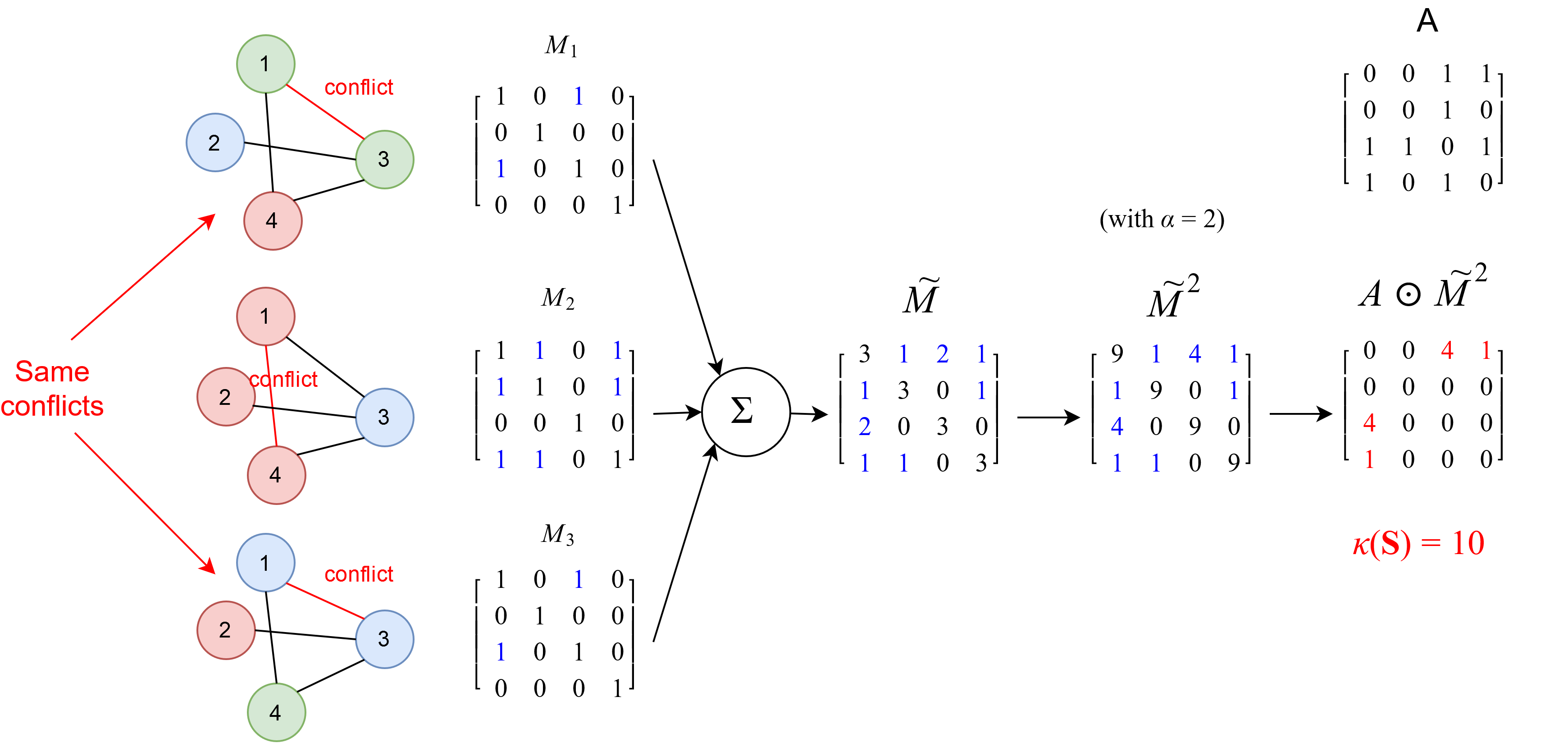}
    \caption{$\kappa(\mathbf{S})=10>6$ for a population of 3 candidate solutions because two solutions share the same conflict on edge $\{1,3\}$.\label{fig:same_conflict}}
    \end{subfigure}
    \caption{\label{fig:kappa_computation} Evaluation of $\kappa(\mathbf{S})$ for two populations of $D=3$ candidate solutions with one conflict in each solution. There are $n=4$ vertices in the graph that are colored with $k=3$ colors. In this example $\alpha$ is set to 2.}
\end{figure}

Figure \ref{fig:kappa_computation} shows the computation of $\kappa(\mathbf{S})$ for a graph with $n=4$ vertices, $k=3$ colors and a population of 3 individuals. In both example \ref{fig:diff_conflict} and \ref{fig:same_conflict}, there is one conflict in each candidate solution. However $\kappa(\mathbf{S}) = 6$ in example   \ref{fig:diff_conflict} as the conflicts are done for different edges, while $\kappa(\mathbf{S}) = 10$ (greater than 6) in example   \ref{fig:same_conflict} because the first and the third candidate solutions make a conflict for the same edge $\{1,3\}$ (even if different colors are used).

\subsubsection{Bonus term for shared correct color assignments \label{bonus-term}}

Then using the same idea the following bonus term $\varpi(\mathbf{S}) = \text{sum}(\bar{A} \odot \tilde{M}^{\circ \beta})$ is computed, where $\bar{A} = J - A$, with $J$ the matrix of ones of size $n\times n$.

\begin{align}
\varpi(\mathbf{S}) &= \sum_{i,j=1}^n (1-a_{i,j}) \tilde{m}_{i,j}^{\beta} 
  \label{eq:bonus}
\end{align}

\noindent  with $\beta$ greater than 1. 

By maximizing this bonus term, the $D$ candidate solutions are encouraged to make the same correct color assignments for any pair of vertices.

\subsubsection{Global loss function}

The global loss function (to be minimized) aggregates the following three criteria: (i) the global coloring fitness $f_{color}(\mathbf{S})$ (equation (\ref{eq:global coloring fitness})), (ii) the optional global equity fitness $f_{eq}(\mathbf{S})$ for the ECP (equation (\ref{eq:equity_fitness_v0})), (iii) the penalization term for shared wrong color assignments $\kappa(\mathbf{S})$ (equation (\ref{eq:diversity})) and (iv) the bonus term for shared correct color assignments $\varpi(\mathbf{S})$ (equation (\ref{eq:bonus})), subtracted in order to maximize it.
 
\begin{equation}\label{eq:global_loss}
    \mathcal{L}(\mathbf{S})(t) = f_{color}(\mathbf{S}) + \nu t f_{eq}(\mathbf{S}) + \lambda t  \kappa(\mathbf{S}) - \mu t \varpi(\mathbf{S})
\end{equation}

where $\nu \geq 0$ ($\nu = 0$ for the GCP since $f_{eq}$ only concerns the ECP); $\lambda > 0$ and $\mu > 0$ are the weighting factors for the penalization term and the bonus term, while $t \in \llbracket 0, maxIter \rrbracket$ is the iteration counter of the algorithm. The introduction of the time dependent parameter $t$ aims to dynamically change the fitness landscape and increase the chance of the algorithm to get out of local minimum traps.  Introducing the equity fitness progressively for the ECP improves the results as it leaves the possibility for the algorithm to find a legal coloring before having to consider the equity constraint too much.

\subsection{Step 4 - Gradient descent}\label{step4}

Gradient descent is a powerful optimization method that has been applied to various applications such as coefficient updates in linear regression and weight adjustments in neural networks \cite{lecun1989backpropagation}, weighted nearest neighbors feature selection \cite{BugataD19}, and dimensionality reduction using similarity projections \cite{SpathisPT19}. In the context of this work, since the set of candidate solutions $\mathbf{S}$ derives from $\mathbf{W}$ (cf. Step 1 above), the problem is  to optimize $\mathbf{W}$ such that $\mathcal{L}(\mathbf{S})(t) = \mathcal{L}(g(\mathbf{W}))(t)$ is minimum with a first order gradient descent method. To make it easier,  $\mathcal{L}(\mathbf{W})$ is written instead of $\mathcal{L}(g(\mathbf{W}))(t)$ hereafter.

Let $\nabla_\mathbf{W} \mathcal{L}(\mathbf{W})$ be the gradient tensor of $\mathcal{L}(\mathbf{W})$ with respect to $\mathbf{W}$  of size $d \times n \times k$ whose element $(d,i,j)$ is $\frac{\partial \mathcal{L}}{\partial w_{d,i,j}}$. Then the entries of the weight tensor $\mathbf{W}$ can be updated at each iteration as

\begin{equation}
   \mathbf{W} \leftarrow \mathbf{W} - \eta \nabla_\mathbf{W} \mathcal{L}(\mathbf{W}),
\end{equation}

\noindent where $\eta > 0$ is a fix learning rate. This kind of first order optimization is classically used to learn neural networks (in stochastic version) and is known to be very efficient to learn models with a huge number of parameters. 

However, in order to allow the method to escape local minima and forget old decisions that were made long ago and are no longer helpful,  a weight smoothing inspired by the work \cite{zhou2016reinforcement} is applied.  Every $nb_{iter}$ iterations all the weights of $\mathbf{W}$ are divided by a constant number $\rho \geq 1$, which can be achieved with a single tensor computation according to

\begin{equation}\label{eq:smoothing}
    \mathbf{W} \leftarrow  \mathbf{W}/\rho.
\end{equation}

This weight smoothing procedure can also be compared to the pheromone evaporation process used in Ant System (AS) \cite{costa1997ants} in the sense that the pheromone guides the ant search toward the most promising solutions, while a decay factor controls the rate at which historic information is lost.

\subsection{Gradient computation with softmax approximation  \label{sec:grad_comput}}

To tackle this real-valued optimization problem by first order gradient descent over $\mathbf{W}$,   $\nabla_\mathbf{W} \mathcal{L}(\mathbf{W})$ needs to be computed. By applying the chain rule \cite{rumelhart1985learning}, it gives for $d, i, j \in \llbracket 1,D \rrbracket \times \llbracket 1,n \rrbracket \times \llbracket 1,k \rrbracket$,

\begin{equation}
\frac{\partial \mathcal{L}}{\partial w_{d,i,j}}  = \sum_{l=1}^{k}   \frac{\partial \mathcal{L}} {\partial s_{d,i,l}} \times \frac{\partial s_{d,i,l}}{\partial w_{d,i,j}}.
\label{eq:chain_rule}
\end{equation}

\subsubsection{Softmax approximation}

Due to the use of the function \textit{argmax} to build $\mathbf{S}$ from $\mathbf{W}$, each partial term  $\frac{\partial s_{d,i,l}}{\partial w_{d,i,j}}$ entering in equation (\ref{eq:chain_rule}) is equal to zero almost everywhere. Therefore,  the \textit{softmax} function will be used as a continuous, differentiable approximation to \textit{argmax}, with informative gradient. 

$\mathbf{S}$ can be approximated by a tensor $\hat{\mathbf{S}}$ of size $D \times n \times k$, where each $\hat{s}_{d,i,j}$ is computed with the elements of $\mathbf{W}$ using the \textit{softmax} function \cite{bridle1990training} as

\begin{equation}\label{eq:softmax}
\hat{s}_{d,i,j} = \frac{e^{ w_{d,i,j}}}{\sum_{l=1}^{k} e^{ w_{d,i,l}}},\ \text{for}\  d,i,j \in \llbracket 1, D \rrbracket \times \llbracket 1, n \rrbracket \times \llbracket 1,k \rrbracket.
\end{equation}

Each entry $\hat{s}_{d,i,j} \in ]0, 1[$ of the tensor $\hat{\mathbf{S}}$ denotes, for each candidate solution $S_d$, a continuous indicator  that the $i$-th vertex $v_i$ selects the $j$-th color group $g_j$ as its group.  In the following  this soft assignment is rewritten for each vertex with a single tensor equation as

\begin{equation}\label{eq:tensor_softmax}
   \hat{S}(\mathbf{W}) = \text{softmax}( \mathbf{W}),
\end{equation}

\noindent where the \textit{softmax} function is applied along the last axis (color group axis) for each candidate solution and each vertex.

For $d,i,j,l$,  $\frac{\partial s_{d,i,l}}{\partial w_{d,i,j}}$ is approximated by $\frac{\partial \hat{s}_{d,i,l}}{\partial w_{d,i,j}} =  \hat{s}_{d,i,l}(\delta_{j,l} - \hat{s}_{d,i,j})$, with $\delta_{j,l}$ the Kronecker symbol equaling 1 if $j=l$ and 0 otherwise. Using this Straight-Through (ST) gradient estimator \cite{bengio2013estimating} of $\frac{\partial s_{d,i,l}}{\partial w_{d,i,j}}$ in equation (\ref{eq:chain_rule}), it gives for $d, i, j \in \llbracket 1,D \rrbracket \times \llbracket 1,n \rrbracket \times \llbracket 1,k \rrbracket$: 

\begin{align}
  \frac{\partial \mathcal{L}}{\partial w_{d,i,j}}  & \approx \sum_{l=1}^{k} \frac{\partial \hat{s}_{d,i,l}}{\partial w_{d,i,j}} \times \frac{\partial \mathcal{L}} {\partial s_{d,i,l}}
  \\ &= \sum_{l=1}^{k}   \hat{s}_{d,i,l}(\delta_{d,j,l} - \hat{s}_{d,i,j}) \times \frac{\partial \mathcal{L}} {\partial s_{d,i,l}}
    \\ &=   \hat{s}_{d,i,j} \frac{\partial \mathcal{L}} {\partial s_{d,i,j}} -   \hat{s}_{d,i,j}\sum_{l=1}^{k} \hat{s}_{d,i,l} \times \frac{\partial \mathcal{L}} {\partial s_{d,i,l}}. \label{eq:single_gradient}
\end{align}
 In the proposed framework however, no sequential loop is used to calculate the gradient of each parameter. Instead a tensor product is simply computed to get the full gradient matrix. This operation can easily be parallellized on GPU devices. With tensor operations, the $D \times n \times k$ equations (\ref{eq:single_gradient}) for $d,i,j \in \llbracket 1,D \rrbracket \times \llbracket 1,n \rrbracket \times \llbracket 1,k \rrbracket$ become then a single equation:

\begin{equation}\label{eq:gradient}
\nabla_\mathbf{W} \mathcal{L} =  \hat{\mathbf{S}} \odot (\nabla_\mathbf{S} \mathcal{L} -  (\hat{\mathbf{S}} \odot \nabla_\mathbf{S} \mathcal{L}) \cdot J),
\end{equation}

\noindent where $\nabla_\mathbf{S} \mathcal{L} = \{\frac{\partial \mathcal{L}}{\partial s_{d,i,j}} \}_{d,i,j}$ is the gradient matrix of size $D \times n \times k$ of $\mathcal{L}$ with respect to $\mathbf{S}$, $J$ is a matrix of 1's of size $k \times k$, $\odot$ is the Hadamard product (element-wise product) and $\cdot$ is the dot product between two tensors. 

\subsubsection{Gradient of $\mathcal{L}$ with respect to $\mathbf{S}$}

In order to compute $\nabla_\mathbf{W} \mathcal{L}(\mathbf{W})$ with equation (\ref{eq:gradient}), it remains to compute $\nabla_\mathbf{S} \mathcal{L}$. According to equation (\ref{eq:global_loss}), it gives

\begin{align*}
    \mathcal{L}(\mathbf{S})(t) &= f_{color}(\mathbf{S}) +  \nu t f_{eq}(\mathbf{S})  + \lambda t  \kappa(\mathbf{S}) - \mu t \varpi(\mathbf{S}) \\
    &= \frac{1}{2}\text{sum}(\mathbf{A} \odot (\mathbf{S} \cdot \mathbf{S}')) +  \nu t f_{eq}(\mathbf{S})  + \lambda t \  \text{sum}(A \odot \tilde{M}^{\circ \alpha}) \\ & \ \ \ \ - \mu t \ \text{sum}(\bar{A} \odot \tilde{M}^{\circ \beta}).
\end{align*}

At iteration $t$, the gradient of this loss with respect to the tensor $\mathbf{S}$ of $D$ solutions is then given by\footnote{Let us note  that the first term $A \cdot S$ in equation (\ref{eq:global_loss_gradient}), the gradient of $f_{color}(\mathbf{S})$, corresponds to the $\gamma$-matrix used in an efficient implementation proposed in \cite{fleurent1996genetic} of the popular TabuCol algorithm for the GCP \cite{hertz1987using}.}

\begin{equation}\label{eq:global_loss_gradient}
  \nabla_{\mathbf{S}} \mathcal{L}(t) = A \cdot \mathbf{S} +  \nu t \nabla_{\mathbf{S}} f_{eq}(\mathbf{S}) +  2\alpha \lambda t (A \odot \tilde{M}^{\circ (\alpha - 1)}) \cdot \mathbf{S}  - 2\beta \mu t (\bar{A} \odot \tilde{M}^{\circ (\beta - 1)}) \cdot \mathbf{S},
\end{equation}

 \noindent   where $\nabla_{\mathbf{S}} f_{eq}(\mathbf{S})$ is the gradient of the additional global equity fitness. For $d, i, j \in \llbracket 1,D \rrbracket \times \llbracket 1,n \rrbracket \times \llbracket 1,k \rrbracket$, each entry  $(d,i,j)$  of $\nabla_{\mathbf{S}} f_{eq}(\mathbf{S})$ is  equal to

\begin{equation} \label{eq:grad_ecp_group}
    \frac{\partial f_{eq}(\mathbf{S})}{\partial s_{d,i,j}}= 
      \begin{cases}
      0 & \text{if}\ \#V_{d,j} = c_1 \ or\  \#V_{d,j} = c_2 \\
      \\
     +1 & \text{if}\ \#V_{d,j} > c_2\\
     \\
    -1 & \text{if}\ \#V_{d,j} < c_1
  \end{cases}
\end{equation}

with $\#V_{d,j} = \sum_{i=1}^n s_{d,i,j}$ the total number of vertices receiving color $j$ for the candidate solution $d$.

\subsection{General TensCol algorithm \label{sec:general_algo}}

The general TensCol algorithm involving the above-mentioned four steps is summarized in Algorithm \ref{algo_Tenscold}.

The algorithm iterates the four composing steps (Sections \ref{step1}-\ref{step4}) until one of the following stopping conditions is reached: 1) the global fitness of one of the $D$ candidate solutions in $\mathbf{S}$ reaches 0 (i.e., a legal $k$-coloring is found); 2) the number of iterations reaches the allowed limit $maxIter$. One can notice from  Algorithm \ref{algo_Tenscold} that it is not necessary to evaluate the penalization term $\kappa(\mathbf{S})$ and bonus terms $\varpi(\mathbf{S})$. It saves computational and memory resources to compute only their gradients with respect to $\mathbf{S}$.

\begin{algorithm}[h]
\caption{$\text{TensCol}$ for graph coloring problems}
\noindent\begin{minipage}{\textwidth}
\begin{algorithmic}\label{algo_Tenscold}
\STATE \ 
\STATE \bf{Input}: \normalfont{graph ${G}$ with adjacency matrix $A$, available colors $\{1,2\dots,k\}$, random seed $r$, and number of maximum allowed iterations $maxIter$;}
\STATE \bf{Output}: \normalfont{a legal $k$-coloring ${G}$ if it is found;}
\STATE \normalfont{$\mathbf{W} \leftarrow \mathbf{W_0}$}  \hfill $/*$ Weight tensor initialization: $w^0_{i,j,d} \sim \mathcal{N}(0,\sigma_0)$ with random seed $r$ $*/$
\STATE \normalfont{$t \leftarrow 0$}
\REPEAT
\STATE
\STATE  \textbf{\# Step 1) Vertex to color assignment}
\STATE \quad i) $\mathbf{S} = \text{one\_hot}(\text{argmax}(\mathbf{W}))$
\STATE
\STATE \textbf{\# Step 2) Fitness evaluation (forward phase)}
\STATE \quad i) $\mathbf{M} = \mathbf{S} \cdot \mathbf{S}'$
\STATE \quad ii) $\mathbf{C} = \mathbf{A} \odot \mathbf{M}$
\STATE \quad iii) Compute the vector of fitness $(f(S_1),..., f(S_D))$
\STATE
\STATE \textbf{\# Step 3 ) Group concentration matrix, penalization and bonus terms evaluation }
\STATE \quad i) $\tilde{M} = \text{sum}(\mathbf{M},1)$ 
\STATE  \quad ii) Compute $\nabla_{\mathbf{S}} \mathcal{L}(t) = A \cdot \mathbf{S} +  \nu t \nabla_{\mathbf{S}} f_{eq}(\mathbf{S}) +  2\alpha \lambda t (A \odot \tilde{M}^{\circ (\alpha - 1)}) \cdot \mathbf{S}  - 2\beta \mu t (\bar{A} \odot \tilde{M}^{\circ (\beta - 1)}) \cdot \mathbf{S}$
\STATE
\STATE \textbf{\# Step 4 ) Weight update by gradient descent }
\STATE \quad  i) Compute $\hat{\mathbf{S}} = \text{softmax}( \mathbf{W})$; 
\STATE \quad  ii) $\nabla_\mathbf{W} \mathcal{L} =  \hat{\mathbf{S}} \odot (\nabla_\mathbf{S} \mathcal{L} -  (\hat{\mathbf{S}} \odot \nabla_\mathbf{S} \mathcal{L}) \cdot J)$ 
\STATE \quad iii) $\mathbf{W} \leftarrow \mathbf{W} - \eta \nabla_{\mathbf{W}} \mathcal{L}$; 
\STATE \quad iv) Every $nb_{iter}$ do $\mathbf{W} \leftarrow \mathbf{W}/\rho$; 
\STATE
\STATE \quad \normalfont{$t \leftarrow  t + 1$}
\STATE
\UNTIL{$\underset{d}{min}f(S_d)=0$ or $t = maxIter$}
 \IF{$\underset{d}{min}f(S_d)=0$}
\RETURN $S_{d^*}$ with $d^* = \underset{d}{argmin}f(S_d)$
\ENDIF
\end{algorithmic}
\end{minipage}
\end{algorithm}

The only stochastic part in TensCol concerns the initialization of the tensor $\mathbf{W}$ whose weights are drawn randomly according to a normal law. This random initialization has an impact on the search. Therefore for practical performance evaluation, the algorithm will be run a lot of times with different random seeds $r$ to initialize the pseudo-random generator of the normal law.

The time complexity of each step of the algorithm is given in Table \ref{table:algo_complexity}. The most costly operation is in $O(Dn^2k)$. Therefore the main factor influencing the execution time of the algorithm is $n$, the size of the graph. This complexity is alleviated in practice thanks to two factors. First, the tensor operations of the algorithm greatly benefit from our implementation with the Pytorch library specially designed for the parallel calculation of tensors on GPU devices. Second, unlike local search based coloring algorithms where each iteration only changes the color of one single vertex in a candidate solution, each iteration of the TensCol algorithm may simultaneously change the colors of many vertices, because each iteration may update the $n*k$ weights of the solution (see the example in the next section). Moreover, our experiments showed that this is quite useful to deal with large graphs, helping to find several record-breaking results for the ECP, as shown in Section \ref{Computational results on the ECP}.

\begin{table}[h]
\centering
\caption{Time complexity of each step of the algorithm.}
\scriptsize
\begin{tabular}{lll} 
\hline
Step & Type of operation & Time complexity\\
\hline
\hline
Step 1 - i) & Finding the color with max weight value for each vertex & $O(Dnk)$ \\
\hline
Step 2 - i) & Tensor dot product between two tensors of size $(D,n,k)$  and $(D,k,n)$ & $O(Dn^2k)$ \\
Step 2 - ii) & Element wise multiplication between two tensors of size $(D,n,n)$ & $O(Dn^2)$ \\
Step 2 - iii) & Fitness vector computation & $O(Dn^2)$ \\
\hline
Step 3 - i) & Sum along first axis of a tensor of size  $(D,n,n)$ & $O(Dn^2)$ \\
Step 3 - ii) & Computation of the gradient of the loss with respect to $S$  & $O(Dn^2k)$ \\
\hline
Step 4 - i) & Tensor softmax operation & $O(Dnk)$ \\
Step 4 - ii) & Computation of the gradient of the loss with respect to $W$ & $O(Dnk)$ \\
Step 4 - iii) & Tensor subtraction & $O(Dnk)$ \\
Step 4 - iii) & Tensor division & $O(Dnk)$ \\
\hline\end{tabular}
\label{table:algo_complexity}
\end{table}

\section{First experimental analysis}  \label{sec:exp_analyzes}

In this section, an illustrative toy example is first presented for the GCP (subsection \ref{toy_GCP_example}) to give intuitive insights on the gradient descent learning. In Subsection \ref{sec:sensitivity},   a sensitivity analysis of the main parameters $\lambda$ and $\mu$ is presented in order to show the importance of the penalization and bonus terms in the global loss function used by the algorithm. Then in subsection \ref{sec:loss_analysis}, the fitness evolution of the algorithm during the search is shown.

\subsection{Toy example for the graph coloring problem \label{toy_GCP_example}}

\begin{figure}[h]
    \centering
    \begin{subfigure}[t]{.45\textwidth}
    \includegraphics[width=1\textwidth]{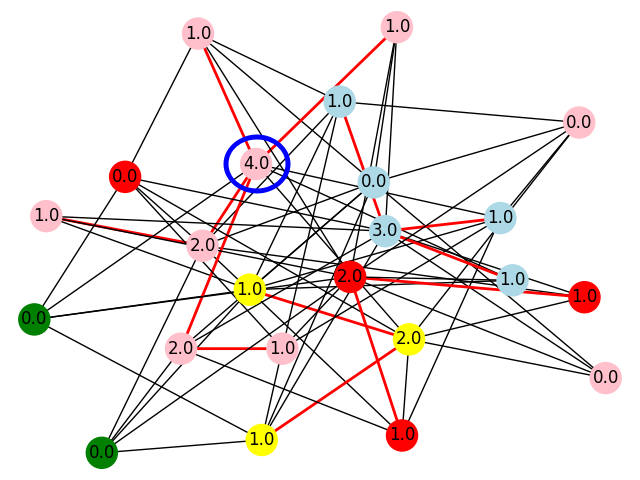}
    \caption{}\label{fig:1gcp}	
    \end{subfigure}
     \begin{subfigure}[t]{.45\textwidth}
    \includegraphics[width=1\textwidth]{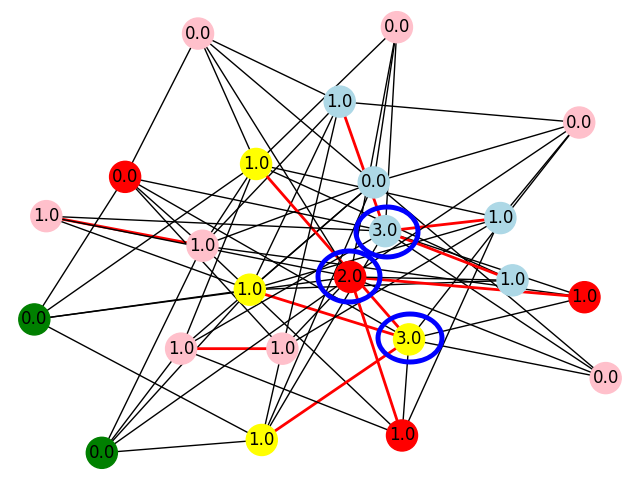}
    \caption{}\label{fig:2gcp}	
    \end{subfigure}
    \begin{subfigure}[t]{.45\textwidth}
    \includegraphics[width=1\textwidth]{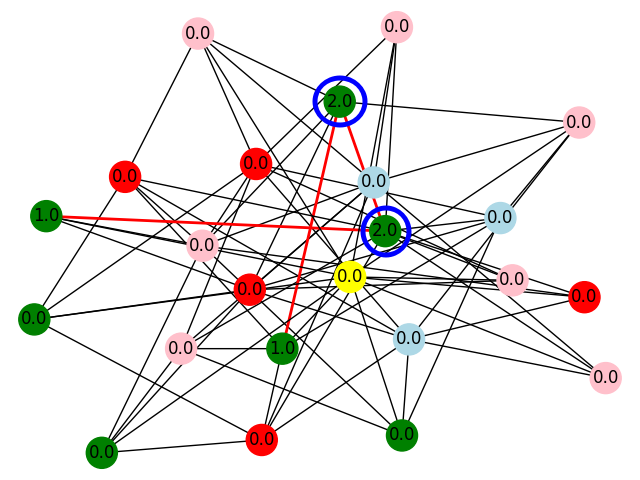}
    \caption{}\label{fig:3gcp}	
     \end{subfigure}
     \begin{subfigure}[t]{.45\textwidth}
    \includegraphics[width=1\textwidth]{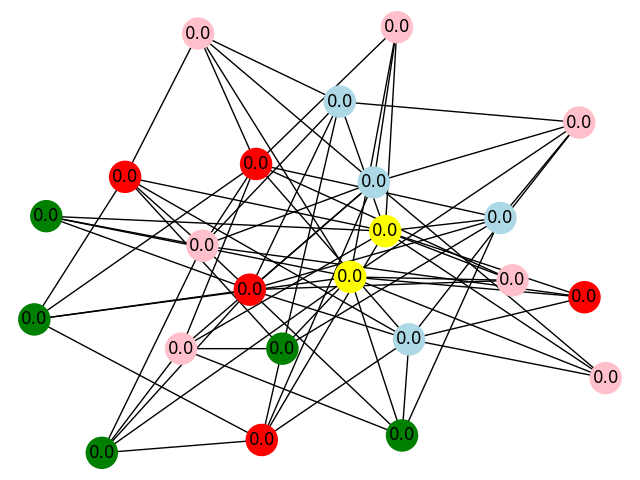}
    \caption{}\label{fig:4gcp}	
    \end{subfigure}
    \caption{\label{fig:GCP_insight} {\bf Last three iterations of $\text{TensCol}$} to color the graph myciel4.col with 5 colors (optimal value). The number on each vertex indicates the gradient of the global fitness score with respect to the selected color for this vertex. Red edges corresponds to conflicting edges. Blue circles highlight the vertices changing their color from one iteration to another. Several vertices can change their colors during the same iteration. Better seen in color.}
\end{figure}

A simplified implementation of $\text{TensCol}$ is used for the GCP, where only one candidate solution is evaluated ($D=1$). It is applied to an easy graph named myciel4.col of the COLOR02 benchmark with 21 vertices and known chromatic number of 5. Figure \ref{fig:GCP_insight} displays the last 3 iterations (over 8) of $\text{TensCol}$ to color myciel4.col with 5 colors and random seed $r=0$. The parameters used in $\text{TensCol}$ are $\lambda=0$, $\mu=0$, $\eta=0.001$, $nb_{iter}=5$ and $\rho =200$.
The number written on each vertex indicates the gradient of the fitness with respect to the selected color for this vertex. It corresponds to the total number of conflicts, i.e., adjacent vertices receiving the same color. Red edges correspond to conflicting edges. Blue circles highlight the vertices that change their color from one iteration to another. As one can see on this figure,  $\text{TensCol}$ tends to change the color of the vertices with the highest gradient values. One can also notice that $\text{TensCol}$ can change the color of more than one vertex at each iteration as the update is done on the full candidate solution matrix $S$ at each iteration. A legal coloring solution $S^*$ is shown in Figure \ref{fig:GCP_insight} (d) with $f(S^*) = 0$.

\subsection{Sensitivity analysis}
\label{sec:sensitivity}

This section investigates the usefulness of the penalization term for shared incorrect group assignments $\kappa(\mathbf{S})$ and the bonus term for shared correct group assignments $\varpi(\mathbf{S})$, which are two important parts of the global loss function (equation \ref{eq:global_loss}) with the two weighting factors $\lambda$ and $\mu$. For this study,  a population of $D=200$ candidate solutions launched in parallel is adopted.

\begin{figure}[!h]
\centering
\includegraphics[width=0.7\textwidth]{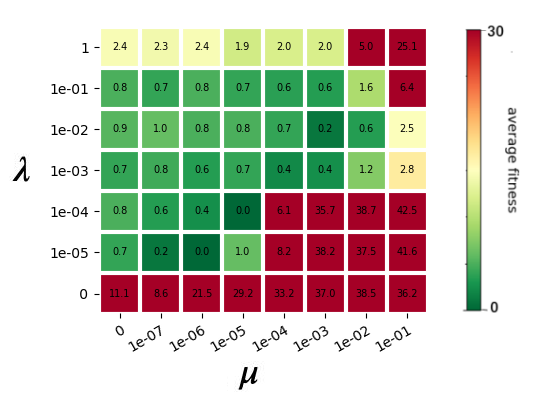}
\caption{Sensitivity analysis of $\lambda$ (weighting factor for the penalization term) and $\mu$ (weighting factor for the bonus term) over the average global fitness of the graph DSJC250.5.col for the GCP. The average fitness is computed from the best global fitness values of 20 different runs, each run performing 200 000 iterations with a new random initialization. Best results are in green.}\label{fig:heatmap}
\end{figure}

The two dimensional heat map presented in Figure \ref{fig:heatmap} shows the average best global fitness obtained by TensCol on the  DSJC250.5.col instance with 250 vertices from  the second competition organized by the Center for Discrete Mathematics and Theoretical Computer Science (DIMACS). This average fitness is computed from 20 different runs, each run performing 200,000 iterations with a new random initialization and new values for $\lambda$ and $\mu$ from the range of 0 to 1.

When $\lambda=0$ and $\mu=0$, meaning that there are no interactions between the $D$ candidate solutions, TensCol fails to obtain a good solution in the allotted time. The parameter that plays the most crucial role is $\lambda$ which is used to penalize shared incorrect group assignments of the $D$ solutions. Best results are obtained for $\lambda \in \llbracket 1e-05,1e-04 \rrbracket$. Furthermore the parameter $\mu$ used with the bonus term to encourage shared correct pairwise assignments, also helps to improve the results when it takes values between 1e-06 and 1e-05. However, its impact is less important than $\lambda$. If $\mu$ is set too high, the results deteriorate as the $D$ candidate solutions are not diverse anymore. Best results with an average fitness equal to 0.0 (corresponding to a success rate of 100\% on the 20 different runs) are obtained for the pairs $(\lambda = 10^{-5}, \mu = 10^{-6})$ and $(\lambda = 10^{-4}, \mu = 10^{-5})$. The first pair of values $(\lambda = 10^{-5}, \mu = 10^{-6})$ will be kept in the comparative benchmark evaluation presented in this work.

\subsection{Evolution of the loss function  \label{sec:loss_analysis}}

To illustrate how the objective converges during the learning and search process, the fitness evolution is studied when applying TensCol to color the random graph R250.5.col (250 vertices) with 65 colors (chromatic number). The algorithm is launched with the parameters $D=200$, $\rho=10$, $nb_{iter}=5$, $\alpha=2.5$, $\lambda = 10^{-5}$, $\beta=1.2$,  $\mu = 10^{-6}$ (for the GCP) and $\nu = 10^{-5}$ (for the ECP).

TensCol reaches a legal coloring (for the GCP) with 17,000 iterations on average (this corresponds to 3.4 million fitness evaluations since 200 candidate solutions are evaluated at the same time). It takes 33 seconds on average to find the solution on a Nvidia RTX2080Ti GPU card with a success rate of 100\%. 

\begin{figure}[!h]
\centering
\includegraphics[width=0.95\textwidth]{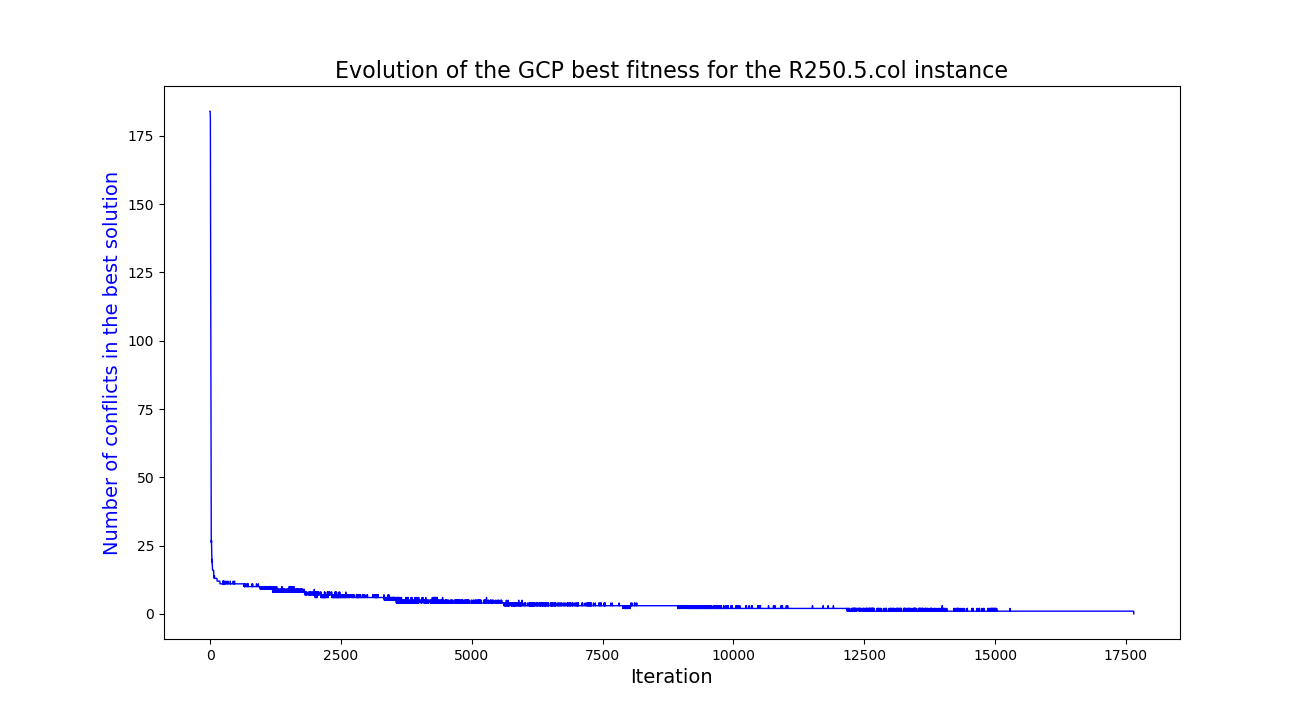}
\caption{Evolution of the number of conflicts obtained by the best solution among the 200 candidate solutions during the search for a legal coloring with 65 colors for the random graph R250.5.col. }\label{fig:loss_gcp}
\end{figure}

\begin{figure}[!h]
\centering
\includegraphics[width=0.95\textwidth]{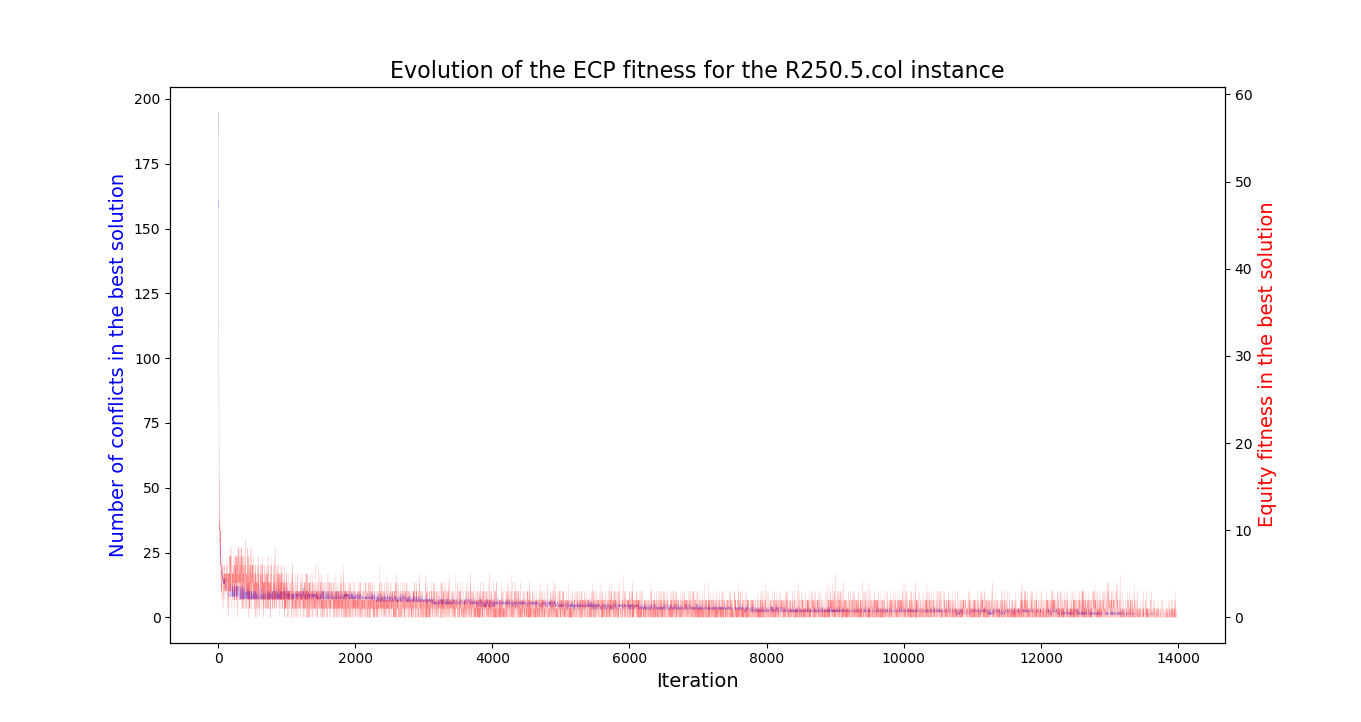}
\caption{Evolution of the number of conflicts (in blue) and the equity fitness (in red) obtained by the best solution among the 200 candidate solutions during the search for a feasible coloring with 65 colors for the random graph R250.5.col. }\label{fig:loss_ecp}
\end{figure}

Figure \ref{fig:loss_gcp} shows that for the GCP, the fitness improves very quickly at the beginning to reach a score of around 15 conflicts and gradually decreases until a legal solution is found. For the ECP, Figure \ref{fig:loss_ecp} indicates oscillations of the equity fitness (in red) as the algorithm quickly moves between equitable and non-equitable colorings during the search before reaching a feasible coloring. Small oscillation peaks appear due to the process of resetting the weights to avoid local optima (cf. Step 4-iv in Algorithm \ref{algo_Tenscold}). 

Finally, the objective convergence profile observed on the graph R250.5.col represents a typical search behavior of the algorithm.

\section{Computational results}  \label{sec:experiments}

This section is dedicated to a computational assessment of the $\text{TensCol}$ algorithm for solving the two target graph coloring problems (i.e., GCP and ECP).   The benchmark instances are first presented (subsection \ref{Benchmark instances and comparison criterion}), followed by the experimental setting of $\text{TensCol}$ for both problems (subsection \ref{Benchmark instances and experimental setting}). Then, the computational results obtained on the GCP are shown in subsection \ref{Computational results on the GCP}, and the results on the ECP are presented in subsection \ref{Computational results on the ECP}. An analysis and discussions of the results obtained on different graph structures are then proposed in subsection \ref{analysis_results}.

\subsection{Benchmark instances and comparison criterion}
\label{Benchmark instances and comparison criterion}

For the GCP, the TensCol algorithm is evaluated on the 36 most difficult benchmark instances from the second DIMACS competition\footnote{Publicly available at \url{ftp://dimacs.rutgers.edu/pub/challenge/graph/benchmarks/color/}} that were used to test graph coloring algorithms in recent studies including the two individual memetic algorithm \cite{moalic2018variations} and the probabilistic learning based algorithm \cite{zhou2018improving}. For the ECP, the same set of 73 benchmark instances used in \cite{sun2017feasible,wang2018tabu} from the second DIMACS and COLOR02 competitions\footnote{Publicly available at \url{https://mat.gsia.cmu.edu/COLOR02/}} is adopted.

Following the common practice to report comparative results in the coloring literature, the assessment focuses on the best solution found by each algorithm corresponding to the smallest number $k$ of colors needed to reach a legal coloring for a graph. It is worth mentioning that for the GCP, no single algorithm in the literature including the most recent algorithms can attain the best-known results for all 36 difficult DIMACS instances. Indeed, even the best performing algorithms miss at least two best-known results. This is understandable given that these instances have been studied for a long time (over some 30 years) and some best-known results have been achieved only by very few algorithms under specific and relaxed conditions (e.g., large run time from several days to one month). Moreover, one notices that for these benchmark graphs, even finding a legal $k$-coloring is a difficult task if $k$ is set to a value slightly above the chromatic number or the current best-known result. In other words, it is extremely difficult and unlikely to find improved solutions for these graphs. As such, an algorithm able to attain (or draw near to) the best-known results for a majority of the benchmark instances can be qualified promising with respect to the current state-of-the-art on graph coloring. Most of the above comments remain valid for the ECP.

Given these comments, one understands that for graph coloring problems (and in general for hard combinatorial optimization), computation time is not the main concern for performance assessment. This is also because the state-of-the-art algorithms (implemented with different programming languages) often report different results for some graphs, which were obtained on various computing platforms with specific stopping conditions (e.g., maximum allowed generations or iterations, maximum allowed fitness evaluations, a cutoff time limit etc). As a result, the timing information, when it is shown, is provided for indicative purposes only.

\subsection{Experimental setting}
\label{Benchmark instances and experimental setting}

The $\text{TensCol}$ algorithm was implemented in Python 3.5 with Pytorch 1.1 library for tensor calculation with CUDA 10.0\footnote{The source code of our algorithm is available at \url{https://github.com/GoudetOlivier/TensCol}.}. It is specifically designed to run on GPU devices. In this work  a Nvidia RTX 2080Ti graphic card with 12 GB memory was used. A Nvidia Tesla P100 with 16 GB memory was used only for the two very large graphs (C2000.5.col and C2000.9.col instances).

For a small graph such as r250.5.col colored with $k=65$ colors for an equitable coloring, when  $\text{TensCol}$ is launched with $D=200$ candidate solutions in parallel, $200*250*65 = 3.2$ millions weights $w_{d,i,j}$ are updated at each iteration. On a Nvidia RTX 2080Ti graphic card, it takes 0.002 seconds per iteration (for 200 candidate solutions evaluated). For a large graph such as C2000.9.col colored with $k=431$ colors for an equitable coloring, there are $200*2000*431 = 172$ millions weights updated at each iteration. On a Nvidia Tesla P100, it takes 0.37 seconds per iteration.

To identify appropriate values for TensCol's parameters,  a grid search is applied on a set of representative instances (DSJC250.5.col, DSJC250.9.col, DSJC500.5.col, DSJC500.9.col,  DSJR500.1c.col, DSJR500.5.col, R250.5.col, R1000.1.col). For the grid search, each parameter is set within a reasonable range ($\lambda \in \{10^{-4}, 10^{-5}, 10^{-6} \}$, $\mu \in \{10^{-5}, 10^{-6}, 10^{-7} \}$ (for the GCP), $\nu \in \{10^{-4}, 10^{-5}, 10^{-6} \}$ (for the ECP), $\alpha \in \{1.5, 2, 2.5 \}$,  $nb_{iter} \in \{5,10,100\}$, $\beta \in \{1.2, 1.5, 2 \}$ and  $\rho \in \{1,2,10,100,$ $200\}$) while keeping the other parameter values with their default values $D=200$, $\sigma_0 = 0.01$ and $\eta = 0.001$. An early stopping iteration criterion of $10^5$ iterations is used to quickly stop the algorithm if the score no longer improves. The grid search allowed us to determine a best baseline parameter setting, given in Table \ref{table:parameters_tenscol}, for the GCP and the ECP, which are kept for all instances, except for the smoothing parameter $\rho$ which is very sensitive to the graph structure. The value of this parameter is thus chosen among 5 different values for each graph. For very difficult random graphs such as DSJC500.5.col and DSJC1000.5.col, the values $\rho=100$ or $\rho=200$ are used in order to frequently smooth the learned weight tensor $\mathbf{W}$ according to equation (\ref{eq:smoothing}). However, for geometric graphs such as R1000.1.col or instances based on register allocation such as fpsol2.i.1.col or mulsol.i.1.col, the value $\rho =1$ is used meaning that no smooth process is applied during the search.

\begin{table}[!h]
\centering
\caption{Parameter setting in $\text{TensCol}$ for the GCP and the ECP}
\begin{scriptsize}
\begin{tabular}{llll} 
Parameter & Description & Value for the GCP & Value for the ECP\\
\hline
$D$ & number of candidate solutions & 200 & 200\\
$\sigma_0$ & standard deviation of initial parameters & 0.01 & 0.01\\
$\eta$ & learning rate & 0.001 & 0.001\\ 
$nb_{iter}$ & smoothing procedures period & 5 & 5\\ 
$\rho$ & smoothing parameter & \{1,2,10,100,200\} & \{1,2,10,100,200\}\\ 
$\alpha$  & exponent for the penalization term & 2.5 & 2.5\\
$\lambda$  & weighting factor for the penalization term & $10^{-5}$ & $10^{-5}$\\
$\beta$ & exponent for the bonus term & 1.2  & 1.2\\
$\mu$ & weighting factor for the bonus term & $10^{-6}$ & $0.0$\\
$\nu$ & weighting factor for the equity constraint & 0.0 & $10^{-5}$\\
\hline
\end{tabular}
\end{scriptsize}
\label{table:parameters_tenscol}
\end{table}

For some particular instances, it is necessary to deviate from this baseline parameter setting in order to reach the state-of-the-art results.  For the random DSJC graphs with a high density (DSJC125.9.col, DSJC250.9.col, DSJC500.9.col and DSJC1000.9.col), the value $\alpha = 1.5$ is adopted instead of $\alpha = 2.5$ in order to prevent the penalization term $\kappa(S)$ from becoming too important (cf. equation (\ref{eq:diversity})). On the contrary, for the random DSJC graphs with a low density (DSJC125.1.col, DSJC250.1.col, DSJC500.1.col and DSJC1000.1.col),  the parameter $\mu$ is set to $10^{-7}$ for the GCP instead of $10^{-6}$ to avoid that the bonus term $\varpi(S)$ (cf. equation (\ref{eq:bonus})) becomes too preponderant in the global loss evaluation. Specific fine tuning of the parameters are also done in the best performing methods such as HEAD \cite{moalic2018variations} and QA \cite{titiloye2011graph}.

For all the tested instances, the stopping criterion used is a maximum allowed number of iterations of $2\times10^6$, which corresponds to a maximum of $4\times10^8$ evaluations of candidate solutions (as there are 200 solutions evaluated at each iteration). For the experiments, each instance was solved 10 times independently (with 10 different random seeds) following the common practice in the literature for graph coloring. 
 
\subsection{Computational results on graph coloring benchmark instances}
\label{Computational results on the GCP}

This section is dedicated to an extensive experimental evaluation of $\text{TensCol}$ on the GCP.  A comparison of $\text{TensCol}$ with 5 best-performing coloring algorithms is proposed (see Section \ref{sec:related_work_gcp} for a  presentation of these algorithms):

\begin{itemize}
    \item the two-phase evolutionary algorithm (MMT, 2008) \cite{malaguti2008metaheuristic} ran on a 2.4 GHz Pentium processor with a cut-off time of 1.6 or 11.1 hours;
    \item the distributed evolutionary quantum annealing algorithm (QA, 2011) \cite{titiloye2011graph} ran on a 3.0 GHz Intel processor with 12 cores with a cut-off time of 5 hours;
    \item the population-based memetic algorithm (MA, 2010) ran on a 3.4 GHz processor with a cut-off time of 5 hours;
    \item the latest hybrid evolutionary algorithm in Duet (HEAD, 2018) \cite{moalic2018variations} ran in parallel  on a 4 cores 3.1 GHz Intel Xeon processor with a cut-off time of at least 3 hours;
	\item the probability learning based local search (PLSCOL, 2018) \cite{zhou2018improving} ran on an Intel Xeon E5-2760 2.8 GHz processor with a cut-off of 5 hours.
\end{itemize}

To the best of our knowledge, these reference algorithms represent the state-of-the-art for the GCP, which cover the current best-known results in the literature for the difficult graphs used in this work.

Detailed results of $\text{TensCol}$ and the reference algorithms on the 36 DIMACS graphs are reported in Tables \ref{table:gcp_results1} and \ref{table:gcp_results2} in the Appendix. Columns 1 and 2 indicate the name and the number of vertices in each instance. Column 3 shows the chromatic number $\chi ({G})$ (if known)\footnote{For some instances such as flat1000\_76\_0 the chromatic number is known by construction of the graph (and equal to 76 in this case), but no heuristic has yet found a legal coloring with the chromatic number.}, while column 4 ($k^*$) gives the best-known results (best upper bound of $\chi ({G})$) ever reported by an algorithm. Columns 5 to 19 show the best results ($k_{best}$) of the reference algorithms with their success rate and computation time (in seconds). The remaining columns report the results of our $\text{TensCol}$ algorithm: the best result ($k_{best}$), the success runs (SR) over 10 runs during which $\text{TensCol}$ attained its best result, and the average computation time (t(s)) for a successful run. A '-' symbol indicates that the corresponding result is not available. As discussed in Section \ref{Benchmark instances and comparison criterion}, computation times were provided for indicative purposes.

As one can observe from Tables \ref{table:gcp_results1} and \ref{table:gcp_results2}, $\text{TensCol}$  always finds the best-known coloring for the instances with less than 200 vertices such as DSJC125.1.col or R125.1. On medium graphs with up to 500 vertices such as DSJC250.5.col or DSJC500.5.col, Tensol is competitive compared to all reference algorithms. For the two large random DSJC1000.5.col and DSJC1000.9.col instances, $\text{TensCol}$ performs better than PLSCOL and MMT, but worse than the two best-performing memetic algorithms MA and HEAD. For the family of flat graphs, MA and HEAD perform the best. But these graphs (especially  flat300\_28\_0 and flat1000\_76\_0 graph) remain difficult for all the algorithms.

$\text{TensCol}$ performs very well on the whole family of 7 geometric graphs by attaining all the best-known results and in particular reports a 234-coloring on the very difficult R1000.5.col graph. This is a remarkable result because this 234-coloring, which was first reported in 2008 by the highly sophisticated MMT algorithm \cite{malaguti2008metaheuristic}, has never been reproduced by another algorithm since that date. $\text{TensCol}$ also obtains an excellent 98-coloring for the large latin\_square\_10 graph, which remains difficult for the reference algorithms except QA which reported the same result as $\text{TensCol}$.

Finally, given the particularity of the GCP benchmark instances as discussed in Section \ref{Benchmark instances and comparison criterion}, it is not suitable to apply statistical tests.

\subsection{Computational results on equitable coloring benchmark instances}
\label{Computational results on the ECP}

This section reports a comparison of $\text{TensCol}$ with 5 most recent state-of-the-art algorithms in the literature for the ECP (more details of these algorithms are given in Section \ref{sec:related_work_ecp}): 

\begin{itemize}
    \item the tabu search algorithm (TabuEqCol, 2014) \cite{diaz2014tabu};
    \item the backtracking based iterated tabu search (BITS, 2015) \cite{lai2015backtracking}; 
    \item the feasible and infeasible search algorithm (FISA, 2017) \cite{sun2017feasible};
    \item the hybrid tabu search (HTS, 2018) \cite{wang2018tabu};
    \item the memetic algorithm (MAECP, 2020) \cite{SunHao2020}.
\end{itemize}

The results of TabuEqCol were obtained on a computer with an Intel i5 CPU750 2.67GHz processor under a cutoff limit of 1h, while the four other reference algorithms were run on an Intel Xeon E5440 2.83 GHz processor (or a comparable processor) with a cutoff time of 2.7h for the instances with upto 500 vertices and 11 hours for larger instances. Given that TabuEqCol is the oldest algorithm among the reference algorithms, it is no surprise that it is dominated by the other algorithms. TabuEqCol is included in this study as it is a baseline reference for the ECP.

Tables \ref{table:ecp_results1} - \ref{table:ecp_results4} show the best results of the six compared algorithms in terms of the smallest number of colors used to color each graph with the equity constraint for the DIMACS and COLOR02 instances. Column 3 displays the overall best-known result $k^*$ of the ECP ever reported in the literature. The next fifteen columns report the best results by the reference algorithms (TabuEqCol, BITS, FISA, HTS and MAECP) with their success rate and computation time (in seconds). The last three columns show the results of our Tensol algorithm. A '-' symbol indicates that the corresponding result is not available. Like for the GCP, the assessment focuses on the quality criterion and timing information is provided for indicative purposes only.

From the results of Tables \ref{table:ecp_results1} - \ref{table:ecp_results4}, one can observe that in terms of attaining the best-known results, $\text{TensCol}$ is the second best performing algorithm after the most recent MAECP algorithm (12 best-known results missed by $\text{TensCol}$ against 8 best-known results missed by MAECP). In particular, for the graphs with up to 500 vertices, $\text{TensCol}$ obtains comparable results with the state-of-the-art algorithms, except for DSJC500.9.col, le450\_15a.col, le450\_15b.col, le450\_25c.col and le450\_25d.col where $\text{TensCol}$ performs worse than some reference algorithms, but it is better on the DSJC500.5.col instance than most competitors. For the wapXXa.col instances, the results obtained by $\text{TensCol}$ are not so good as the other algorithms.

On the contrary, $\text{TensCol}$ excels on large graphs between 900 and 2000 vertices. Remarkably, it established 8 new best-known records (new upper bounds summarized in Table \ref{table:ecp}) by finding legal and equitable colorings with fewer colors compared to the current best-known $k^*$ values. As one can observe in Table \ref{table:ecp}, $\text{TensCol}$ significantly improves the best-known upper bound for three large graphs with 37 colors less for C2000.9.col, 9 colors less for C2000.5.col and 8 colors less for R1000.5.col, while the improvement for the 6 other cases goes from -1 to -3 colors. These new solutions are available at \url{https://github.com/GoudetOlivier/TensCol}.

\begin{table}[!h]
\centering
\caption{New record coloring results found by $\text{TensCol}$ for 8 large benchmark graphs. Some improvements are very important with a gain of 8, 9 and 37 colors.  \label{tab:new_col8 dooring}}
\scriptsize
\begin{tabular}{lcccc} 
\hline
Instance & $|V|$  & previous best-known $k^*$ &  new best-known $k^*$ & Improvement\\
\hline
DSJR500.5.col & 500  & 124 & 122 & -2\\
DSJC1000.5.col & 1000  & 95 & 92 & -3\\
R1000.5.col & 1000  & 247 & 239 & -8\\
flat1000\_60.0.col & 1000  & 93 & 92 & -1\\
flat1000\_76.0.col & 1000  & 93 & 91 & -2\\
latin\_square\_10.col & 900  & 103 & 102 & -1\\
C2000.5.col & 2000  & 183 & 172 & -9\\
C2000.9.col & 2000  & 468 & 431 & -37\\
\hline\end{tabular}
\label{table:ecp}
\end{table}

\subsection{Analysis of the results and discussions}
\label{analysis_results}

This section discusses the reasons behind the results presented in the comparative study of the last two subsections.

\subsubsection{Analysis of the graph coloring results}

One can notice in Tables \ref{table:gcp_results1} and \ref{table:gcp_results2} that compared to most reference algorithms, $\text{TensCol}$ competes favorably on the \textit{geometric} graphs such as the large R1000.5.col graph by finding 234-colorings. Meanwhile, compared to some reference algorithms, $\text{TensCol}$ is less competitive on some \textit{random} graphs such as DSJC500.5.col and DSJC1000.5.col.

\begin{figure}[H]
    \centering
    \includegraphics[width=0.65\textwidth]{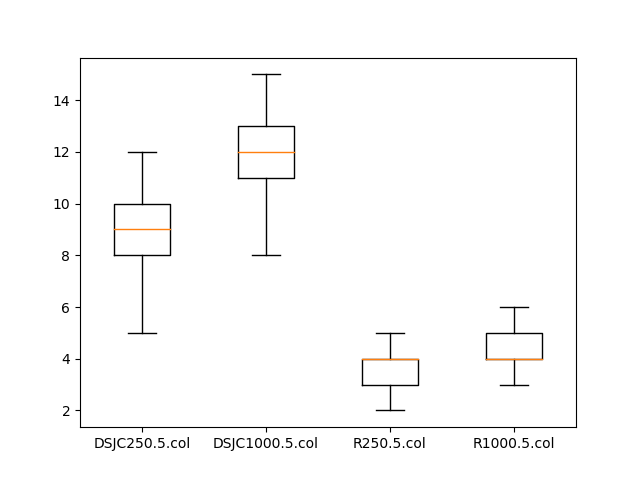}
    \caption{\label{fig:boxplot} Boxplot of the size distribution of the color groups in the solutions for graphs DSJC250.5.col, DSJC1000.5.col, R250.5.col and R1000.5.col colored by $\text{TensCol}$ with 28, 84, 65 and 234 colors, respectively.}
\end{figure}

To better understand these results, one can look at the structure of the solutions  obtained with the $\text{TensCol}$ algorithm for the random graphs DSJC250.5.col ($k=28$) and DSJC1000.5.col ($k=84$), and for the geometric graphs R250.5.col ($k=65$) and R1000.5.col ($k=234$). All these graphs have an edge density of 0.5. However the distributions of the sizes of the color groups in the solutions for these two types of graphs are very different as shown in Figure \ref{fig:boxplot}. For the random graphs DSJC250.5.col and DSJC1000.5.col, there are important variations in color group  sizes with many (12 to 15) shared vertices, while the color groups in the solutions of the geometric graphs are more homogeneous, mainly with small color groups of size 4 or 5\footnote{Let us note that the fact that large color groups are not part of any optimal 234-coloring for the R1000.5.col instance was also observed in \cite{wu2012coloring}.}.
 
$\text{TensCol}$ does not explore (large) shared color groups during the search. Instead, it inherits and learns \textit{pairwise} information via the computation of the \textit{group concentration matrix} $\tilde{M}$ with equation (\ref{eq:tildeM}). In this matrix $\tilde{M}$, each entry $(i,j)$ indicates the number of candidate solutions with vertices $v_i$ and $v_j$ being assigned to the same color group. This pairwise information sharing contributes to satisfactorily coloring geometric graphs. On the other hand, it seems to be a disadvantage when dealing with large random graphics (since it does not inherit large shared color groups). 

This \textit{pairwise} information sharing strategy is in sharp contrast to the reference algorithms that rely on recombination of large color groups (such as MA and HEAD with the Greedy Partition Crossover (GPX) of \cite{galinier1999hybrid}). These algorithms are expected to perform well when large shared color groups are parts of a solution (this is the case for random graphs such as DSJC250.5.col and DSJC1000.5.col) and may fail if this condition is not met as for the geometric graphs such as R250.5.col and R1000.5.col. Indeed, an inspection of elite candidate solutions (with conflicts) from the HEAD algorithm for coloring R1000.5.col with $k = 234$ showed that the shared color groups inherited from parent solutions include at least 7 vertices, while the largest color group in legal 234-colorings has 6 vertices only. This indicates that memetic algorithms with the GPX crossover (such as MA and HEAD) can be misguided by the recombination, thus leading to mediocre results for this geometric graph.

These discussions lead to the conclusion that $\text{TensCol}$ can be considered as a complementary approach with respect to the existing approaches, especially, the memetic approach.

\subsubsection{Analysis of the equitable coloring results}

$\text{TensCol}$ has attained a very good performance on many instances for the ECP, including 8 new record-breaking results. $\text{TensCol}$ seems powerful in finding solutions with homogeneous sizes of color groups for the GCP, this may explain its good results for the ECP. Indeed, given the equity constraint, the color graphs of a solution for the ECP are necessarily of homogeneous sizes.

However $\text{TensCol}$ also reported bad results (see Table \ref{table:ecp_results2}) for some specific instances, in particular for the wapXXa.col graphs.  Two reasons can be advanced to explain this counter-performance.

First, most of the wapXXa.col graphs are very large with more than 2000 vertices. Processing such a large graph by  
$\text{TensCol}$ with $D=200$ solutions evaluated in parallel requires more than 16 Go on the GPU devices, surpassing the memory limit of the computer used in this work. Therefore it was necessary to reduce the number of candidate solutions evaluated at the same time to $D=20$, which impacts negatively the performance of the $\text{TensCol}$ algorithm.

Second, most of the wapXXa.col instances present a very low density of edges (between 2\% and 8\%) with a high number of vertices. Therefore the number $k$ of colors required to color these graphs is very low compared to the number $n$ of vertices (for the instance wap01a.col, the number of vertices is $n=2368$ and the best coloring found in the literature only requires $k=42$ colors). It was empirically noticed that the gradient descent used by TensCol is less informative in this case because the number of weights updated ($n\times k$)  at each iteration of the algorithm is reduced. In fact, $\text{TensCol}$ benefits less from one of its main advantages which is to encode information on the possible colors for each vertex. In the extreme case of binary group assignment, with only two weights per vertex,  our weight learning strategy is expected to be less effective compared to other approaches. This is consistent with the presented results that TensCol is more effective for large graphs with a high number of groups such as R1000.5, C2000.5 and C2000.9 colored with 239, 172 and 431 colors (see Tables \ref{table:ecp_results1} and \ref{table:ecp_results3}).

\section{Other applications of the proposed method} \label{Discussion}

In this section  the generality of the proposed approach is discussed by showing how it can be conveniently applied to solve other graph problems such as multicoloring, minimum sum coloring and $k$-partitioning. 

In the multicoloring problem \cite{halldorsson2004multicoloring}, each vertex $v_i \in V$ is allocated a weight $c_i \in \{1,2,...\}$ and the task is to assign $c_i$ different colors to each vertex $v_i$ such that adjacent vertices have no colors in common and the number of colors used is minimal. For this problem, as $c_i$ different colors have to be assigned to each vertex, a modification has to be done in the vertex to color assignment procedure of the $\text{TensCol}$ algorithm (Step 1). Instead of selecting for the $i$-th vertex the color group with the maximum weight in $w_i$, a set of $c_i$ color groups corresponding to the  $c_i$ maximum values of $w_i$ could be selected.

In the minimum sum coloring problem \cite{jin2017algorithms}, a weight $z_l$ is associated with each color and the objective is to find a legal coloring which minimizes the sum of the total color cost. The $\text{TensCol}$ algorithm could be used by replacing the equity fitness $f_{eq}(S_d)$ in Algorithm \ref{algo_Tenscold} for each candidate solution by the additive fitness $f_{mscp}(S_d) = \sum_{i=1}^n  \sum_{l=1}^k s_{d,i,l} \times z_{l}$ corresponding to the sum of the total color cost of a candidate solution $S_d$.

For the popular $k$-partitioning problem \cite{fjallstrom1998algorithms},  the same TensCol algorithm could be used except that $M$ would be replaced by $\bar{M}=J-M$, where $J$ is the matrix of ones of size $n \times n$. Then, the fitness of a candidate solution for the $k$-partitioning problem can be evaluaed by $f(S_d)= \frac{1}{2}\text{sum}(A \odot \bar{M})$, corresponding to the number of cut edges between partitions. For the $k$-partitioning variant with a balancing constraint \cite{andreev2006balanced}, the same additional \textit{equity fitness} $f_{eq}(S_d)$ of a candidate  solution $S_d$  given by equation (\ref{eq:equity_fitness_v0}) for the ECP could be added to the fitness function of the graph partitioning problem.

\section{Conclusion \label{sec:conclusion}}

A population-based gradient descent weight learning approach is presented in this paper, as well as a practical implementation on GPU devices to solve graph coloring problems. This approach has the particular feature of being general and can be applied to different coloring (and other) problems with a suitable global loss function. This is in sharp contrast to existing approaches for graph coloring, which are specific to a problem. The proposed approach formulates the computation of a solution as a weight tensor optimization problem which is solved by a first order gradient descent. As such it can take advantage of GPU accelerated training to deal with multiple parallel solution evaluations.

The proposed approach was assessed on the well-known graph coloring problem and the related equitable coloring problem. The results of extensive experimental evaluations on popular DIMACS and COLOR02 challenge benchmark graphs showed that $\text{TensCol}$ competes globally well with the best algorithms for both problems with an unified approach. Overall, even if $\text{TensCol}$ does not find new record results for the GCP (this rarely happened since 2012), it provides good results in particular for large geometric graphs (such as R1000.5.col or latin\_square\_10.col) which are particularly difficult for the best performing memetic algorithms. For the ECP, our method discovered 8 improved best results (new upper bounds) for large geometric graphs and random graphs, while the improvements for several graphs are very important in terms of the number of gained colors. Given the novelty of the approach and its promising results reached, this work can serve as the starting point for new research to further advance the state-of-the-art on solving graph coloring and other related problems as discussed in the last section.

A detailed analysis of the achieved results reveals however three main limitations of the proposed algorithm. First, the $\text{TensCol}$ algorithm can be sensitive to its parameters for specific instances, especially regarding the $\rho$ and $\lambda$ parameters. An adaptive strategy could be employed to adjust these parameters depending on the graph encountered \cite{AutomaticParameterTuning2020}. Second, due to the memory capacity of the GPU devices used for our experiments, $\text{TensCol}$ has troubles to deal with very large graphs with more than 2000 vertices. This situation will change because GPU devices with much more memory are expected to become increasingly affordable in the coming years. Third, $\text{TensCol}$ does not learn information about large shared color groups like the best performing memetic algorithms do with the GPX-like crossover. This partially explains the mediocre results of $\text{TensCol}$ on some random instances of the GCP. It would be interesting to extend the $\text{TensCol}$ algorithm to take into account group information. However, this feature also proves to be an advantage to deal with large geometric graphs for the GCP and for the ECP where the equity constraint is imposed. $\text{TensCol}$ can therefore be seen as a complementary approach with respect to the best existing approaches based on the {memetic} framework for graph coloring problems.

Other future works could be performed, but are not limited to the following. It would be interesting to test the proposed approach on other graph problems including those mentioned in Section \ref{Discussion}. Furthermore, the optimization process of the learned weight tensor can be improved by using more sophisticated methods such as second order gradient descent with Hessian tensor, adaptive learning rate or momentum \cite{bertsekas2015convex}. To learn more complex dependencies between the group assignments, it would be worth investigating advanced strategies to replace the weight tensor of candidate solutions by a product of tensors or even a deep generative neural network \cite{bengio2014deep}.

Finally, the two coloring problems studied in this work are general models that can be used to formulate a number of real applications, such as scheduling and timetabling (see the book \cite{Lewis16} for more application examples). The algorithm presented in this work could be adopted by researchers and practitioners to solve such real problems. The availability of the source code of our algorithm will further facilitate such applications.

\section*{Declaration of competing interest}
The authors declare that they have no known competing financial interests or personal relationships that could have appeared to influence the work reported in this paper.

\section*{Acknowledgments}
We are grateful to the reviewers for their useful comments and suggestions which helped us to significantly improve the paper.
This work was granted access to the HPC resources of IDRIS under the allocation 2019-AD011011161 made by GENCI (French national supercomputing facility).

\bibliography{biblio}
\bibliographystyle{plain}

\section*{Appendix - Comparative results of $\text{TensCol}$ with state-of-the-art algorithms on the GCP and the ECP}

 \begin{landscape}
\begin{table}[!h]
\centering
\caption{Comparative results of $\text{TensCol}$ with state-of-the-art algorithms on the set of 36 most difficult DIMACS graphs for the GCP (1/2). Numbers in bold indicate that the best result $k_{\text{best}}$ found by the algorithm attains the overall best-known result in the literature $k^*$ or the chromatic number.}
\scriptsize
\begin{tabular}{l|l|ll|lll|lll|lll|lll|lll|lll} 
   \hline
   &  &  & & \multicolumn{3}{|c|}{PLSCOL} & \multicolumn{3}{|c|}{MMT} & \multicolumn{3}{|c|}{MA} & \multicolumn{3}{|c|}{QA}  & \multicolumn{3}{|c|}{HEAD} & \multicolumn{3}{|c}{$\text{TensCol}$}\\
   \hline
  Instance & $|V|$ & $\chi(G)$ & $k^*$ & $k_{\text{best}}$ & SR & t(s)& $k_{\text{best}}$ & SR & t(s)& $k_{\text{best}}$ & SR & t(s)& $k_{\text{best}}$ & SR & t(s) & $k_{\text{best}}$ & SR & t(s)  \\
    \hline
    DSJC125.1 & 125 & 5 & 5   & \textbf{5} & 10/10 &  $<60$ & \textbf{5} & - &  21 & \textbf{5} & 10/10 &  60 & - & - & - & \textbf{5} & 20/20 &  $<1$ & \textbf{5} & 10/10 &  40 \\
    DSJC125.5 & 125 & 17 & 17  & \textbf{17} & 10/10 &  $<60$ & \textbf{17} & - &  122 & \textbf{17} & 10/10 &  180 & - & - & - & \textbf{17} &  20/20 & $<1$ &\textbf{17} & 10/10  & 68  \\
    DSJC125.9 & 125 & 44 & 44   & \textbf{44} & 10/10 &  $<60$ & \textbf{44} & - &  121 & \textbf{44} & 10/10 &  240 & - & - & - & \textbf{44} &  20/20 & $<1$ & \textbf{44} & 10/10  & 22 \\       
    DSJC250.1 & 250 & ? & 8   & \textbf{8} & 10/10 &  $<60$ & \textbf{8} & - &  21 & \textbf{8} & 10/10 &  120 & - & - & - & \textbf{8} &  20/20 & $<1$ & \textbf{8} & 10/10  & 95 \\  
    DSJC250.5 & 250 & ? & 28  & \textbf{28} & 10/10 &  4 & \textbf{28} & - &  117 & \textbf{28} & 20/20 &  $<60$ & \textbf{28} & 10/10 & 8 & \textbf{28} & 20/20 & 0.6 &\textbf{28} & 10/10  & 199 \\  
    DSJC250.9 & 250 & ? & 72  & \textbf{72} & 10/10 &  $<60$ & \textbf{72} & - &  89 & \textbf{72} & 10/10 &  180 &  - & - & - & \textbf{72} &  20/20 & 1.2 & \textbf{72} & 10/10  & 87 \\  
    DSJC500.1 & 500 & ? & 12   & \textbf{12} & 7/10 &  43 & \textbf{12} & - &  210 & \textbf{12} & 20/20 &  $<60$ & \textbf{12} & 10/10 & 82 & \textbf{12} & 20/20 & 6 &\textbf{12} & 10/10  & 1098 \\ 
    DSJC500.5 & 500 & ? & 47   & 48 & 3/10 &  1786 & 48 & - &  388 & 48 & 20/20 &  1320 & 48 & 10/10 & 494 & \textbf{47} & 2/10000 & 48 & 48 & 5/10  & 7807\\
    DSJC500.9 & 500 & ? & 126   & \textbf{126} & 10/10 &  747 & 127 & - &  433 & \textbf{126} & 20/20 &  5700 & \textbf{126} & 10/10 & 1198 & \textbf{126} & 13/20 & 72 &\textbf{126} & 6/10  & 18433 \\
    DSJC1000.1 & 1000 & ? & 20   & \textbf{20} & 1/10 &  3694 & \textbf{20} & - &  260 & \textbf{20} & 16/20 &  6480 & \textbf{20}  & 9/10  & 2842 & \textbf{20} & 20/20 & 12 &\textbf{20} & 10/10  & 10225 \\
    DSJC1000.5 & 1000 & ? & 82   & 87 & 10/10 &  1419 & 84 & - &  8407 & 83 & 20/20 &  2820 & 83 & 9/10 & 12773 & \textbf{82} & 3/20 & 2880 & 84 & 9/10  & 32495 \\  
    DSJC1000.9 & 1000 & ? & 222  & 223 & 5/10 &  12094 & 225 & - &  3234 & 223 & 18/20 &  9000 & \textbf{222} & 2/10 & 13740 & \textbf{222} & 2/20 & 5160 & 224 & 6/10  & 58084 \\  
    DSJR500.1 & 500 & 12 & 12  & \textbf{12} & 10/10 &  $<60$ & \textbf{12} & - &  25 & \textbf{12} & 10/10  & 240 & - & - &  - & - & - &  - & \textbf{12} & 10/10  & 7 \\ 
    DSJR500.1c & 500 & ? & 85 & \textbf{85}  & 10/10 &  386 & \textbf{85} & - &  88 & \textbf{85} & 20/20  & 300 & \textbf{85} & 10/10 & 525 & \textbf{85} & 1/20 & 12 & \textbf{85} & 10/10  & 298 \\ 
    DSJR500.5 & 500 & ? & 122  & 126  & 8/10 &  1860 & \textbf{122} & - &  163 & \textbf{122} & 11/20  & 6900 & \textbf{122} & 2/10 & 370 & - & - & - & \textbf{122} & 10/10  & 4310\\ 
    flat300\_26\_0 & 300 & 26  &  26 & \textbf{26}  & 10/10 &  195 & \textbf{26}& - &  36  & \textbf{26} & 20/20  & 240 & - & - & - & - & - & - & \textbf{26} & 10/10  & 176\\  
    flat300\_28\_0 & 300 & 28 & 28  &  30  & 10/10 &  233 & 31 & - &  212 & 29 & 15/20  & 7680 & 31 & 10/10  & 19 & 31 & 20/20 & 1.2 & 31 & 10/10  & 586\\  
    flat1000\_76\_0 & 1000 & 76 & 81  & 86  & 1/10 &  5301 & 83 & - &  7325 & 82 & 20/20  & 4080 & 82 & 7/10  & 9802 &  \textbf{81} & 3/20 & 3600 & 83 & 3/10  & 34349 \\  
    latin\_square\_10 & 900 & ? & 97   & 99  & 8/10 &  2005 & 101 & - &  5156 & 99 & 5/20  & 9480 & 98 & 10/10  & 1449 &  - & - & - & 98 &  10/10 & 28925\\
    \hline
\end{tabular}
\label{table:gcp_results1}
\end{table}
\end{landscape}

 \begin{landscape}
\begin{table}[!h]
\centering
\caption{Comparative results of $\text{TensCol}$ with state-of-the-art algorithms on the set of 36 most difficult DIMACS graphs for the GCP (2/2). Numbers in bold indicate that the best result $k_{\text{best}}$ found by the algorithm attains the overall best-known result in the literature $k^*$ or the chromatic number.}
\scriptsize
\begin{tabular}{l|l|ll|lll|lll|lll|lll|lll|lll} 
   \hline
   &  &  & & \multicolumn{3}{|c|}{PLSCOL} &  \multicolumn{3}{|c|}{MMT} & \multicolumn{3}{|c|}{MA} & \multicolumn{3}{|c|}{QA}  & \multicolumn{3}{|c|}{HEAD} & \multicolumn{3}{|c}{$\text{TensCol}$}\\
   \hline
  Instance & $|V|$ & $\chi(G)$ & $k^*$ & $k_{\text{best}}$ & SR & t(s)& $k_{\text{best}}$ & SR & t(s)& $k_{\text{best}}$ & SR & t(s)& $k_{\text{best}}$ & SR & t(s)& $k_{\text{best}}$ & SR & t(s) & $k_{\text{best}}$ & SR & t(s)  \\
    \hline
    le450\_15a & 450 & 15 & 15  & \textbf{15} & 10/10 &  $<60$ & \textbf{15} & - &  $<1$ & \textbf{15}  & 10/10  & 120 & - & - & - &  \textbf{15}  & 20/20 & $<1$ & \textbf{15} & 10/10  & 333\\ 
    le450\_15b & 450 & 15 & 15  & \textbf{15} & 10/10 &  $<60$ & \textbf{15} & - &  $<1$ & \textbf{15}  & 10/10  & 120 & - & - & - &  \textbf{15}  & 20/20 & $<1$ & \textbf{15} & 10/10  & 333 \\  
    le450\_15c & 450 & 15 & 15  & \textbf{15}  & 7/10 &  1718 & \textbf{15}& - & 3 & \textbf{15} & 20/20  & 180  & - & - & - & \textbf{15}  & 20/20 & 1.2 &\textbf{15} & 10/10  & 507 \\ 
    le450\_15d & 450 & 15 & 15  & \textbf{15}  & 3/10 &  2499 & \textbf{15} & - &  4 & \textbf{15} & 20/20  & 300 & - & - & - & \textbf{15}  & 1/20 & 1.8 & \textbf{15} & 10/10  & 301 \\    
    le450\_25a & 450 & 25 & 25  & \textbf{25} & 10/10 &  $<60$ &  - & - & - & \textbf{25} & 10/10  & 120 & - & - & - &  \textbf{25} & 20/20 & $<1$ &\textbf{25} & 10/10  & 87 \\  
    le450\_25b & 450 & 25 & 25  & \textbf{25} & 10/10 &  $<60$ & - & - & - & \textbf{25} & 10/10  & 180 & - & - & - & \textbf{25} & 20/20 & $<1$ & \textbf{25} & 10/10  & 12 \\ 
    le450\_25c & 450 & 25 & 25  & \textbf{25}  & 10/10 &  1296 & \textbf{25} & - & 1321 & \textbf{25} & 20/20  & 900 & - & - & - & \textbf{25} & 20/20 & 1800 &\textbf{25} & 10/10  & 19680 \\  
    le450\_25d & 450 & 25 & 25  & \textbf{25}  & 10/10 &  1704 & \textbf{25} & - &  436 & \textbf{25} & 20/20  & 600 & - & - & - & \textbf{25} & 20/20 & 5400 &\textbf{25} & 10/10  & 9549\\   
    R125.1 & 125 & 5 & 5  & \textbf{5} & 10/10 &  $<60$  & \textbf{5} & - &  $<1$ & \textbf{5} & 10/10  & 120 & - & - & - & - & - & -  & \textbf{5} & 10/10  & 0.03 \\  
    R125.5 & 125 & 36 & 36 & \textbf{36} & 10/10 &  $<60$ &  \textbf{36}  & - &  21 &  \textbf{36} & - &  $<1$ & \textbf{36} & 10/10  & 60  & - & - & - & \textbf{36} & 10/10  & 6.2 \\ 
    R250.1 & 250 & 8 & 8  & \textbf{8} & 10/10 &  $<60$ & \textbf{8}  & - &  $<1$ & \textbf{8} & 10/10  & 300 & - & - & - & - & - & - &\textbf{8} & 10/10  & 0.04 \\
    R250.5 & 250 & 65 & 65  & 66  & 10/10 &  705 & \textbf{65}  & - &  64 & \textbf{65} & 20/20  & 240 & \textbf{65} & 9/10  & 168 & \textbf{65} & 1/20 & 780 & \textbf{65} & 10/10  & 33 \\
    R1000.1 & 1000 & ? & 20 & \textbf{20} & 10/10 &  $<60$ & \textbf{20}  & - &  37 & \textbf{20}  & 10/10  & 120 & - & - & - & - & - & - & \textbf{20} & 10/10  & 15 \\ 
    R1000.1c & 1000 & ? & 98  &   \textbf{98}   & 10/10 &  256 & \textbf{98}  & - &  518& \textbf{98} & 20/20  & 480 & \textbf{98} & 10/10  & 287 & \textbf{98} & 3/20 & 12 & \textbf{98} & 10/10  & 4707 \\ 
    R1000.5 & 1000 & ? & 234  &  254   & 4/10 &  7818 & \textbf{234}  & - &  753 & 245 & 13/20  & 16560 & 238 & 3/10  & 9511 & 245 & 20/20 & 14640 & \textbf{234} & 2/10  & 23692 \\ 
    \hline
\end{tabular}
\label{table:gcp_results2}
\end{table}
\end{landscape}


\begin{landscape}
\begin{table}[!h]
\centering
\scriptsize
\caption{\footnotesize{Comparative results of $\text{TensCol}$ with state-of-the-art algorithms on the 73 benchmark ECP instances (1/4). Numbers in bold indicate that the best result $k_{\text{best}}$ found by the algorithm is equal to the overall best-known result $k^*$ in the literature. A star in column $k_{\text{best}}$ for $\text{TensCol}$ indicates that a new best coloring of the ECP has been found.}}
\begin{tabular}{l|l|l|lll|lll|lll|lll|lll|lll} 
   \hline
   &  &  & \multicolumn{3}{|c|}{TabuEqCol} & \multicolumn{3}{|c|}{BITS} & \multicolumn{3}{|c|}{FISA} & \multicolumn{3}{|c|}{HTS} & \multicolumn{3}{|c|}{MAECP} & \multicolumn{3}{|c}{$\text{TensCol}$} \\
   \hline
  Instance & $|V|$  & $k^*$ & $k_{\text{best}}$ & SR & t(s) & $k_{\text{best}}$ & SR & t(s) & $k_{\text{best}}$ & SR & t(s) & $k_{\text{best}}$ & SR & t(s) & $k_{\text{best}}$ & SR & t(s) & $k_{\text{best}}$ & SR & t(s) \\
    \hline
    DSJC125.1 & 125  & 5 & \textbf{5} & - &  0.8 & \textbf{5} & 20/20 &  0.96 & \textbf{5} & 20/20 &  0.62 & \textbf{5} & 20/20 & 15.7 & \textbf{5} & 20/20 & 1.26 & $\mathbf{5}$ & 10/10 & 50 \\
    DSJC125.5 & 125  & 9  & 18 & - &  788 & \textbf{17} & 10/20 & 5169 & \textbf{17} & 20/20 &  428 & \textbf{17} & 20/20 & 563 & \textbf{17} & 19/20 & 2432 & $\mathbf{17}$ & 10/10 & 124\\
    DSJC125.9 & 125  & 44 & 45 & - &  0.4 & \textbf{44} & 20/20 & 0.16 & \textbf{44} & 20/20 &  0.09 & \textbf{44} & 20/20 & 0.3 & \textbf{44} & 20/20 & 2.8 & $\mathbf{44}$ & 10/10 & 22\\
    DSJC250.1 & 250  & 8 & \textbf{8} & - &  32 & \textbf{8} & 20/20 & 5.5 & \textbf{8} & 20/20 &  3.62 & \textbf{8} & 20/20 & 427 & \textbf{8} & 20/20 & 4.8 &  $\mathbf{8}$ & 10/10 & 97\\
    DSJC250.5 & 250  & 29 & 32 & - &  69 & 30 & 1/20 & 3265 & \textbf{29} & 13/20 &  5236 & \textbf{29} & 20/20 & 4584 & \textbf{29} & 20/20 & 1093  &  $\mathbf{29}$ & 10/10 & 312\\
    DSJC250.9 & 250  & 72 & 83 & - &  1.2 & \textbf{72} & 20/20 & 1180 & \textbf{72} & 20/20 &  892 & \textbf{72} & 20/20 & 1835 & \textbf{72} & 18/20 & 2540 & $\mathbf{72}$ & 9/10 &  1285\\
    DSJC500.1 & 500  & 13 & \textbf{13} & - &  33 & \textbf{13} & 20/20 & 6.96 & \textbf{13} & 20/20 &  3.57 & \textbf{13} & 20/20 & 148 & \textbf{13} & 20/20 & 112.67 & $\mathbf{13}$ & 10/10 & 168\\
    DSJC500.5 & 500  & 51 & 63 & - &  11 & 56 & 1/20 & 484 & 52 & 1/20 &  8197 & 52 & 20/20 & 2098 & \textbf{51} & 1/20 & 20784 & $\mathbf{51}$ & 10/10 & 3793\\
    DSJC500.9 & 500  & 128 & 182 & - &  0.7 & 129 & 2/20 & 3556 & 130 & 3/20 &  6269 & 129 &  7/20 & 8926 & \textbf{128}  & 2/20 & 16170 & 129 & 4/10 & 13537\\
    DSJR500.1 & 500  & 12 & \textbf{12} & - &  0.0 & \textbf{12} & 20/20 & 0.58 & \textbf{12} & 20/20 &  0.38 & \textbf{12} &  20/20 & 2.53 & \textbf{12} & 20/20 & 13.9 & $\mathbf{12}$ & 10/10 &  490\\  
    DSJR500.5 & 500  & 124 & 133 & - &  0.1 & 126 & 14/20 & 3947 & 126 & 10/20 &  4459 & 125 &  7/20 & 8179 & 124 & 1/20 & 13266 & \textbf{122}* & 10/10 & 5021\\  
    DSJC1000.1 & 1000  & 21 & 22 & - &  500 &  \textbf{21} & 1/20 & 3605 & \textbf{21} & 20/20 &  1866 & \textbf{21} &  1/20 & 4809 & \textbf{21} & 17/20 & 1365 & \textbf{21} & 9/10 &  1757\\ 
    DSJC1000.5 & 1000 & 95 & 112 & - &  2261 & 103  & 3/20 & 18079 &  95 & 2/20 &  15698 & 95 & 6/20 & 13394 & 95   & 3/20 & 36321 & \textbf{92}* & 10/10 & 12430\\ 
    DSJC1000.9 & 1000  & 251  & 329 & - &  20 & 252  & 1/20 & 4065 & 252 & 16/20 &  2240 & \textbf{251} & 20/20 & 3564 & \textbf{251} & 20/20 & 963 & \textbf{251} & 10/10 & 20862 \\
    R125.1 & 125  & 5 & - & - & - & \textbf{5} & 20/20 & 0.01 & \textbf{5} & 20/20 &  0 & \textbf{5} & 20/20 & 0.06 & \textbf{5}  & 20/20 & 0.37 & \textbf{5} & 10/10 & 0.07 \\
    R125.5 & 36  & 36 & - & - & - & \textbf{36} & 20/20 & 0.39 & \textbf{36} & 20/20 &  0.67 & \textbf{36} & 20/20 & 3.41 & \textbf{36} & 20/20 & 1.76 &  \textbf{36} & 10/10 & 29\\ 
    R250.1 & 250  & 8  & - & - & - & \textbf{8} & 20/20 & 0.01 & \textbf{8} & 20/20 &  0 & \textbf{8} & 20/20 & 0.11 & \textbf{8} & 20/20 & 0.24  & \textbf{8} & 10/10 & 0.13 \\ 
    R250.5 & 250  & 65 & - & - & - & 66 & 7/20 & 6275 & 66 & 2/20 &  3041 & \textbf{65} & 2/20 & 9778 & \textbf{65} & 3/20 & 11291  &  \textbf{65} & 10/10 & 40\\ 
    R1000.1 & 1000  & 20 & - & - & - & \textbf{20} & 20/20 & 3.09 & \textbf{20} & 20/20 &  2.24 & \textbf{20} & 20/20 & 678 & \textbf{20} & 20/20 & 21  & \textbf{20} & 10/10 & 4426\\ 
    R1000.5 & 1000  & 247 & - & - & - & 250 & 12/20 & 10723 & 250 & 11/20 &  11564 & 249 & 19/20 & 17817 & 247 & 3/20 & 11291 &  \textbf{239}* & 9/10 & 53187\\ 
\hline
\end{tabular}
\label{table:ecp_results1}
\end{table}
\end{landscape}

\begin{landscape}
\begin{table}[!h]
\centering
\caption{Comparative results of TensCol with state-of-the-art algorithms on the 73 benchmark ECP instances (2/4). Numbers in bold indicate that the best result $k_{\text{best}}$ found by the algorithm is equal to the overall best-known result $k^*$ in the literature. A star in column $k_{\text{best}}$ for TensCol indicates that a new best coloring of the ECP has been found.}
\scriptsize
\begin{tabular}{l|l|l|lll|lll|lll|lll|lll|lll} 
   \hline
   &  &  & \multicolumn{3}{|c|}{TabuEqCol} & \multicolumn{3}{|c|}{BITS} & \multicolumn{3}{|c|}{FISA} & \multicolumn{3}{|c|}{HTS} & \multicolumn{3}{|c|}{MAECP} & \multicolumn{3}{|c}{$\text{TensCol}$} \\
   \hline
     Instance & $|V|$  & $k^*$ & $k_{\text{best}}$ & SR & t(s) & $k_{\text{best}}$ & SR & t(s) & $k_{\text{best}}$ & SR & t(s) & $k_{\text{best}}$ & SR & t(s) & $k_{\text{best}}$ & SR & t(s) & $k_{\text{best}}$ & SR & t(s) \\
    \hline
      le450\_5a & 450  & 5 & - & - & - & \textbf{5} & 20/20 & 46 & \textbf{5} & 20/20 &  30.2 & \textbf{5} & 20/20 & 332 & \textbf{5} & 20/20 & 38.4 &  \textbf{5} & 10/10 & 15\\ 
    le450\_5b & 450  & 5 & 7 & - & 7.2 & \textbf{5} & 20/20 & 74 & \textbf{5} & 20/20 &  44.29 & \textbf{5} & 20/20 & 364 & \textbf{5} & 20/20 & 64.7 &  \textbf{5} & 10/10 & 22\\ 
    le450\_5c & 450  & 5 & - & - & - & \textbf{5} & 20/20 & 1877 & \textbf{5} & 20/20 &  16.39 & \textbf{5} & 20/20 & 143 & \textbf{5} & 20/20 & 17.5 &  \textbf{5} & 10/10 & 4\\ 
    le450\_5d & 450  & 5 & 8 & - & 15 & \textbf{5} & 20/20 & 2231 & \textbf{5} & 20/20 &  14.07 & \textbf{5} & 20/20 & 373 & \textbf{5} & 20/20 & 16.1 &  \textbf{5} & 10/10 & 3\\ 
    le450\_15a & 450  & 15 & - & - & - & \textbf{15} & 20/20 & 4.44 & \textbf{15} & 20/20 & 2.99 & \textbf{15} & 20/20 & 17 & \textbf{15} & 20/20 & 7.4 & 18 & 10/10 & 4252\\ 
    le450\_15b & 450  & 15 & \textbf{15} & - & 107 & \textbf{15} & 20/20 & 4.16 & \textbf{15} & 20/20 &  2.41 & \textbf{15} & 20/20 & 31664 & \textbf{15} & 20/20 & 8.0  &  18 & 10/10 & 3346\\ 
    le450\_15c & 450  & 15 & - & - & - & \textbf{15} & 18/20 & 410 & \textbf{15} & 16/20 &  553 & \textbf{15} & 20/20 & 96 & \textbf{15} & 20/20 & 1351 &  \textbf{15} & 10/10 & 120\\
    le450\_15d & 450  & 15 & 16 & - &  599 & \textbf{15} & 6/20 & 629 & \textbf{15} & 3/20 &  638 & \textbf{15} & 20/20 & 58 & \textbf{15} & 13/20 & 4992 &  \textbf{15} & 10/10 & 178\\ 
    le450\_25a & 450  & 25 & - & - & - &  \textbf{25} & 20/20 & 0.72 & \textbf{25} & 20/20 &  0.41 & \textbf{25} & 20/20 & 4.8 & \textbf{25} & 20/20 & 20.2 & \textbf{25} & 10/10 & 6343\\ 
    le450\_25b & 450  & 25 & \textbf{25} & - &  0.0 & \textbf{25} & 20/20 & 0.78 & \textbf{25} & 20/20 &  0.46 & \textbf{25} & 20/20 & 7.3 & \textbf{25} & 20/20 & 28.1 & \textbf{25} & 10/10 & 2411\\
    le450\_25c & 450 & 26 & - & - & - & \textbf{26} & 20/20 & 16.5 & \textbf{26} & 20/20 &  86.9 & \textbf{26} & 20/20 & 11.8 & \textbf{26} & 20/20 & 148 &  31 & 2/10 & 46129\\
    le450\_25d & 450  & 26 & 27 & - &  29 & \textbf{26} & 20/20 & 14.08 & \textbf{26} & 20/20 & 95.9 & \textbf{26} & 20/20 & 6.3 & \textbf{26} & 20/20 & 61 &  31 & 10/10 & 33419\\
    wap01a & 2368  & 42 & 46 & - &  15 & \textbf{42} & 8/20 & 4183 & \textbf{42} & 1/20 &  4545 & \textbf{42} & 20/20 & 700 & \textbf{42} & 20/20 & 10304 & 46 & 2/10 & 1891 \\
    wap02a & 2464  & 41 & 44 & - &  83 & \textbf{41} & 4/20 & 6829 & \textbf{41} & 2/20 &  2538 & \textbf{41}  & 20/20 & 2233 & \textbf{41} & 20/20 & 14295 & 47 & 8/10 & 1657 \\
    wap03a & 4730  & 44 & 50 & - &  464 & 45 & 19/20 & 11267 & 45 & 6/20 &  20202 & 45  & 18/20 & 6150 & \textbf{44} & 2/20 & 34445 & 51 & 2/10 & 5522 \\
    wap04a & 5231  & 43 & - & - & - & 44 & 17/20 & 11345 & 44 & 11/20 &  15614 & 44 & 12/20 & 12845 & \textbf{43} & 2/20 & 33286 & 51 & 6/10 & 2376 \\
    wap05a & 905 & 50 & -& -& -  & \textbf{50} & 20/20 & 8.46 & \textbf{50} & 20/20 &  99.26 & \textbf{50} & 20/20 & 0.18 & \textbf{50} & 20/20 & 10983 & \textbf{50} & 10/10 & 417  \\
   wap06a & 947  & 41 & -& -& - & \textbf{41} & 6/20 & 6892 & \textbf{41} & 1/20 & 9340 & \textbf{41} & 20/20 & 1206 & \textbf{41}  & 19/20 & 13739 & 45 & 8/10 & 413 \\
    wap07a & 1809  & 42 & - & -& -& 43 & 19/20 & 718 & 43 & 19/20 &  4077 & \textbf{42} & 12/20 & 2591 & \textbf{42} & 5/20 & 11304 & 49 & 10/10 & 995 \\
   wap08a & 1870  & 42 & - & -& - & 43 & 19/20 & 951 & 43 & 10/20 &  4872 & \textbf{42} & 6/20 & 3801 & \textbf{42} & 18/20 & 13821 & 47 & 2/10 & 1171 \\
\hline
\end{tabular}
\label{table:ecp_results2}
\end{table}
\end{landscape}

\begin{landscape}
\begin{table}[!h]
\centering
\caption{Comparative results of TensCol with state-of-the-art algorithms on the 73 benchmark ECP instances (3/4). Numbers in bold indicate that the best result $k_{\text{best}}$ found by the algorithm is equal to the overall best-known result $k^*$ in the literature. A star in column $k_{\text{best}}$ for TensCol indicates that a new best coloring of the ECP has been found.}
\scriptsize
\begin{tabular}{l|l|l|lll|lll|lll|lll|lll|lll} 
   \hline
   &  &  & \multicolumn{3}{|c|}{TabuEqCol} & \multicolumn{3}{|c|}{BITS} & \multicolumn{3}{|c|}{FISA} & \multicolumn{3}{|c|}{HTS} & \multicolumn{3}{|c|}{MAECP} & \multicolumn{3}{|c}{$\text{TensCol}$} \\
   \hline
     Instance & $|V|$  & $k^*$ & $k_{\text{best}}$ & SR & t(s) & $k_{\text{best}}$ & SR & t(s) & $k_{\text{best}}$ & SR & t(s) & $k_{\text{best}}$ & SR & t(s) & $k_{\text{best}}$ & SR & t(s) & $k_{\text{best}}$ & SR & t(s) \\
    \hline
flat300\_28.0 & 300  & 32 & 36 & - &  3222 & 34 & 6/20 & 4407 & \textbf{32} & 1/20 &  4910 & 33 & 20/20 & 3801 & \textbf{32} & 7/20 & 5209 & \textbf{32} & 10/10 & 1583\\
flat1000\_50.0 & 1000  & 92 & - & - & - & 101 & 1/20 & 9206 & 94 & 6/20 &  17321 & \textbf{92} & 1/20 & 18035 & 93   & 2/20 & 16779 & \textbf{92} & 7/10 & 5566\\
flat1000\_60.0 & 1000  & 93 & - & - & - & 102 & 5/20 & 10201 & 94 & 5/20 &  10489 & 94 & 8/20 & 16263 & 93 & 3/20 & 14715 &  \textbf{92}* & 10/10 & 5905\\
flat1000\_76.0 & 1000  & 93 & 112 & - &  1572  & 102 & 3/20 & 13063 & 94 & 2/20 &  15246 & 93 & 5/20 & 14307 & 93 & 2/20 & 24103 & \textbf{91}* & 3/10 & 53553\\
latin\_square\_10 & 900  & 103 & 130 & - &  1301 & 115 & 1/20 & 17859 & 104 & 10/20 &  12666 & 107 & 3/20 & 18055 & \textbf{103} & 1/20 & 32403 &  \textbf{103} & 8/10 & 37233\\
 & & & & & & & & & & &  &  &  &  &  &  &  &  \textbf{102}* & 1/10 & 14484\\
C2000.5 & 2000  & 183 & - & - & - & 201 & 7/20 & 4808 & 183 & 13/20 &  19702 & 188 & 1/20 & 19915 & 183 & 20/20 & 4555 & \textbf{172}* & 8/10 & 111884 \\
C2000.9 & 2000  & 468 & - & - & - & 502 & 11/20 & 7772 & 493 & 2/20 &  21164 & 501 & 20/20 & 3952 & 468 & 1/20 & 36966 &  \textbf{431}* & 4/10 & 109243\\
mulsol.i.1 & 197  & 49 & 50 & - &  0.0 & \textbf{49} & 20/20 & 14.82 & \textbf{49} & 20/20 &  44.34 & \textbf{49}  & 20/20 & 0.53 & \textbf{49} & 20/20 & 7.9  &  \textbf{49} & 10/10 & 301\\
mulsol.i.2 & 188  & 36 & 48 & - &  0.1 & \textbf{36} & 13/20 & 3633 & \textbf{36} & 2/20 &  1914 & \textbf{36} & 20/20 & 15 & \textbf{36} & 20/20 & 188 &   \textbf{36} & 3/10 & 2877\\
fpsol2.i.1 & 496  & 65 & 78 & - &  0.1 & \textbf{65} & 20/20 & 8302 & \textbf{65} & 20/20 &  1723 & \textbf{65} & 20/20 & 12.61 & \textbf{65} & 20/20 & 777 &  \textbf{65} & 10/10 & 3033\\
fpsol2.i.2 & 451  & 47 & 60 & - &  0.0 & \textbf{47} & 20/20 & 976 & \textbf{47} & 17/20 &  2357 & \textbf{47} & 20/20 & 6.64 & \textbf{47} & 20/20 & 4983  &   \textbf{47} & 10/10 & 1725\\
fpsol2.i.3 & 425  & 55 & 79 & - &  0.0 & \textbf{55} & 20/20 & 729 &  \textbf{55}& 20/20 &  1310  & \textbf{55} & 20/20 & 3.77 & \textbf{55} & 20/20 & 746 &  \textbf{55} & 10/10 & 135 \\
inithx.i.1 & 864  & 54 & 66 & - &  0.1 &  \textbf{54} & 20/20 & 1468 & \textbf{54} & 7/20 &  3356 & \textbf{54} & 20/20 & 47.3 & \textbf{54}  & 20/20 & 1708 &   \textbf{54} & 9/10 & 11620\\
inithx.i.2 & 645  & 35 & 93 & - &  7.2 & 36 & 13/20 & 12412 & 36 & 5/20 &  3275 & \textbf{35} & 20/20 & 207.4 & \textbf{35} & 20/20 & 4106 &   \textbf{35} & 10/10 & 3039\\
inithx.i.3 & 621  & 36 & - & - & - & 37 & 11/20 & 9214 & 37& 4/20 &  2891  & \textbf{36}  & 20/20 & 5256 & \textbf{36}  & 20/20 & 6529 &   \textbf{36} & 10/10 & 4943\\
zeroin.i.1 & 211  & 49 & 51 & - &  0.0 & \textbf{49} & 20/20 & 1367 & \textbf{49}& 8/20 &  1089  & \textbf{49} & 20/20 & 2.24 & \textbf{49} & 7/20 & 5749 &   \textbf{49} & 10/10 & 449\\
zeroin.i.2 & 211  & 36 & 51 & - &  0.0 &  \textbf{36} & 20/20 &  97 & \textbf{36} & 20/20 &  124 & \textbf{36} & 20/20 & 0.7 & \textbf{36} & 20/20 & 66.7 &   \textbf{36} & 10/10 & 66\\
zeroin.i.3 & 206 & 36 & 49 & - &  0.0 & \textbf{36} & 20/20 &  109 & \textbf{36} & 20/20 &  129 & \textbf{36} & 20/20 & 1.14 & \textbf{36} & 20/20 & 166 &  \textbf{36} & 10/10 & 65\\
myciel6 & 95  & 7 & \textbf{7} & - &  0.0 & \textbf{7} & 20/20 &  0.0 & \textbf{7} & 20/20 &  0 & \textbf{7} & 20/20 & 0.01 & \textbf{7} & 20/20 & 0.56 &   \textbf{7} & 10/10 & 5\\
myciel7 & 191  & 8 & \textbf{8} & - &  0.1 & \textbf{8} & 20/20 &  0.02 & \textbf{8} & 20/20 &  0.02 & \textbf{8} & 20/20 & 0.17 & \textbf{8}  & 20/20 & 1.81 &   \textbf{8} & 10/10 & 9\\
\hline
\end{tabular}
\label{table:ecp_results3}
\end{table}
\end{landscape}

\begin{landscape}
\begin{table}[!h]
\centering
\caption{Comparative results of TensCol with state-of-the-art algorithms on the 73 benchmark ECP instances (4/4). Numbers in bold indicate that the best result $k_{\text{best}}$ found by the algorithm is equal to the overall best-known result $k^*$ in the literature. A star in column $k_{\text{best}}$ for TensCol indicates that a new best coloring of the ECP has been found.}
\scriptsize
\begin{tabular}{l|l|l|lll|lll|lll|lll|lll|lll} 
   \hline
   &  &  & \multicolumn{3}{|c|}{TabuEqCol} & \multicolumn{3}{|c|}{BITS} & \multicolumn{3}{|c|}{FISA} & \multicolumn{3}{|c|}{HTS} & \multicolumn{3}{|c|}{MAECP} & \multicolumn{3}{|c}{$\text{TensCol}$} \\
   \hline
     Instance & $|V|$  & $k^*$ & $k_{\text{best}}$ & SR & t(s) & $k_{\text{best}}$ & SR & t(s) & $k_{\text{best}}$ & SR & t(s) & $k_{\text{best}}$ & SR & t(s) & $k_{\text{best}}$ & SR & t(s) & $k_{\text{best}}$ & SR & t(s) \\
    \hline
4\_FullIns\_3 & 114  & 7 & 7 & - &  0.1 & \textbf{7} & 20/20 &  0.0 & \textbf{7} & 20/20 &  0 & \textbf{7} & 20/20 & 0.01 & \textbf{7} & 20/20 & 0 &  \textbf{7} & 10/10 & 10\\
4\_FullIns\_4 & 690  & 8 & \textbf{8} & - &  0.4 & \textbf{8} & 20/20 &  0.23 & \textbf{8} & 20/20 &  0.12 & \textbf{8} & 20/20 & 3.98 & \textbf{8} & 20/20 & 2.36 &   \textbf{8} & 10/10 & 26\\
4\_FullIns\_5 & 4146  & 9 & \textbf{9} & - &  254 & \textbf{9} & 20/20 &  54.49 & \textbf{9} & 20/20 &  0.12 & \textbf{9} & 14/20 & 57.7 & \textbf{9} & 20/20 & 72.5 &  \textbf{9} & 10/10 & 5191\\
1\_Insertions\_6 & 607  & 7 & \textbf{7} & - &  0.2 & \textbf{7} & 20/20 &  0.34 & \textbf{7} & 20/20 &  0.17 & \textbf{7} & 20/20 & 0.38 & \textbf{7} & 20/20 & 4.61 &   \textbf{7} & 10/10 & 20\\
2\_Insertions\_5 & 597  & 6 & \textbf{6} & - &  0.0 & \textbf{6} & 20/20 &  0.11 & \textbf{6} & 20/20 &  0.06 & \textbf{6} & 20/20 & 0.96  & \textbf{6} & 20/20 & 1.22  &   \textbf{6} & 10/10 & 17\\
3\_Insertions\_5 & 1406  & 6 & \textbf{6} & - &  1.2 & \textbf{6} & 20/20 &  0.57 & \textbf{6} & 20/20 &  0.35 & \textbf{6} & 20/20 & 21.94  & \textbf{6}  & 20/20 & 2.77 &   \textbf{6} & 10/10 & 641\\
school1 & 385  & 15 & \textbf{15} & - &  12 & \textbf{15} & 20/20 &  1.3 & \textbf{15} & 20/20 &  0.93 & \textbf{15} & 20/20 & 0.72  & \textbf{15}  & 20/20 & 9.17 &   \textbf{15} & 10/10 & 16\\
school1\_nsh & 352  & 14 & 14 &  -  & \textbf{14} & 20/20 &  2.63 & \textbf{14} & 20/20 &  1.77 & \textbf{14} & 20/20 & 56.1  & \textbf{14} & \textbf{14} & 20/20 & 17.3  &   \textbf{14} & 8/10 & 1181\\
qg.order40 & 1600  & 40 & \textbf{40} & - &  47 & \textbf{40} & 20/20 &  4.73 & \textbf{40} & 20/20 &  3.44 & \textbf{40} & 20/20 & 4291  & \textbf{40} & 20/20 & 1627 &   \textbf{40} & 10/10 & 44\\
qg.order60 & 3600  & 60 & \textbf{60} & - &  267 & \textbf{60} & 20/20 &  21.57 & \textbf{60} & 20/20 &  14.53 & \textbf{60} & 20/20 & 64.3  & \textbf{60} & 20/20 & 1015 &  \textbf{60} & 10/10 & 655\\
ash331GPIA & 662  & 4 & \textbf{4} & - &  2 & \textbf{4} & 20/20 &  2.02 & \textbf{4} & 20/20 &  0.78 & \textbf{4}& 20/20 & 5.39  & \textbf{4} & 20/20 & 43.8 &   \textbf{4} & 10/10 & 554 \\
ash608GPIA & 1216  & 4  & \textbf{4} & - &  12 & \textbf{4} & 20/20 &  0.5 & \textbf{4}  & 20/20 &  0.25 & \textbf{4} & 20/20 & 854 & \textbf{4}  & 20/20 & 0.44 &  \textbf{4}  & 9/10 & 749\\
ash958GPIA & 1916  & 4  & \textbf{4} & - &  41 & \textbf{4}& 20/20 &  23.31 & \textbf{4} & 20/20 &  10.89 & \textbf{4} & 20/20 & 11.95 & \textbf{4} & 20/20 & 166 &  \textbf{4} & 4/10 & 2819 \\
\hline
\end{tabular}
\label{table:ecp_results4}
\end{table}
\end{landscape}

\end{document}